# High Dimensional Multivariate Regression and Precision Matrix Estimation via Nonconvex Optimization

Jinghui Chen* and Quanquan Gu†

**Abstract**

We propose a nonconvex estimator for joint multivariate regression and precision matrix estimation in the high dimensional regime, under sparsity constraints. A gradient descent algorithm with hard thresholding is developed to solve the nonconvex estimator, and it attains a linear rate of convergence to the true regression coefficients and precision matrix simultaneously, up to the statistical error. Compared with existing methods along this line of research, which have little theoretical guarantee, the proposed algorithm not only is computationally much more efficient with provable convergence guarantee, but also attains the optimal finite sample statistical rate up to a logarithmic factor. Thorough experiments on both synthetic and real datasets back up our theory.

## 1 Introduction

Multiple output multivariate regression refers to the regression analysis of the relationships between several output vectors (responses) and several input vectors (predictors). The goal is to simultaneously exploit the relationship between the input and output, as well as their respective intrinsic dependencies. Applications of this regression model arise in a wide variety of fields such as economics (Srivastava and Giles, 1987) and genomics (Kim et al., 2009; Lee and Liu, 2012; Wang, 2013). The standard multiple output multivariate regression model is described as follows:

$$\mathbf{y} = \mathbf{W}^{*\top}\mathbf{x} + \boldsymbol{\epsilon}, \tag{1.1}$$

where $\mathbf{x} \in \mathbb{R}^d$ is the predictor vector, $\mathbf{y} \in \mathbb{R}^m$ is the response vector, $\mathbf{W}^* \in \mathbb{R}^{d \times m}$ is the unknown but fixed regression coefficient matrix, and $\boldsymbol{\epsilon} \in \mathbb{R}^m$ is the error vector. We assume $\mathbf{x}$ and $\boldsymbol{\epsilon}$ are independent from each other and the error vector $\boldsymbol{\epsilon}$ follows a multivariate normal distribution with zero mean and covariance $\boldsymbol{\Sigma}^*$. In model (1.1), the corresponding precision matrix $\boldsymbol{\Omega}^* = \boldsymbol{\Sigma}^{*-1}$ characterizes the conditional dependency structure among the responses (Wang, 2013). More specifically, $\Omega^*_{ij} = 0$ implies that $i$-th response and $j$-th response are conditionally independent given the covariates and other response variables.

*Department of Systems and Information Engineering, University of Virginia, Charlottesville, VA 22904, USA; e-mail: jc4zg@virginia.edu

†Department of Systems and Information Engineering, University of Virginia, Charlottesville, VA 22904, USA; e-mail: qg5w@virginia.edu

Given $n$ independent and identically distributed observations $\{(\mathbf{x}_i, \mathbf{y}_i)\}_{i=1}^n$ from model (1.1), our goal is to estimate the unknown regression coefficient matrix $\mathbf{W}^*$. A large body of existing work studies the ordinary least squares (OLS) estimator, and its penalized variants (Obozinski et al., 2011; Lounici et al., 2009; Negahban and Wainwright, 2011). However, they often assume the error vector follows from an isotropic multivariate normal distribution with zero mean and identity matrix covariance, which leads to a suboptimal statistical rate when $\mathbf{\Omega}^*$ is not the identity matrix. To address this problem, one can simultaneously estimate the unknown parameters $\mathbf{W}^*$ and $\mathbf{\Omega}^*$ based on the following joint likelihood function of the model in (1.1):

$$\prod_{i=1}^n (2\pi)^{-m/2}|\mathbf{\Omega}|^{1/2} \exp\Big[-\frac{1}{2}\big((\mathbf{y}_i - \mathbf{W}\mathbf{x}_i)^\top \mathbf{\Omega}(\mathbf{y}_i - \mathbf{W}\mathbf{x}_i)\big)\Big],$$

where $\mathbf{\Omega} = \mathbf{\Sigma}^{-1} \in \mathbb{R}^{m \times m}$ is the precision matrix, and $|\mathbf{\Omega}|$ is the determinant of $\mathbf{\Omega}$. The negative log-likelihood function of the above model is equivalent to the following sample loss function up to some constant:

$$f_n(\mathbf{W}, \mathbf{\Omega}) = -\log|\mathbf{\Omega}| + \frac{1}{n} \operatorname{tr}\big[(\mathbf{Y} - \mathbf{X}\mathbf{W})\mathbf{\Omega}(\mathbf{Y} - \mathbf{X}\mathbf{W})^\top\big], \quad (1.2)$$

where $\mathbf{X} = [\mathbf{x}_1, \ldots, \mathbf{x}_n]^\top \in \mathbb{R}^{n \times d}$ is the design matrix, $\mathbf{Y} = [\mathbf{y}_1, \ldots, \mathbf{y}_n]^\top \in \mathbb{R}^{n \times m}$ is the response matrix, and $\mathbf{E} = [\boldsymbol{\epsilon}_1, \ldots, \boldsymbol{\epsilon}_n]^\top \in \mathbb{R}^{n \times m}$ is the error matrix. In many applications such as economics and genomics, the number of parameters $dm + \binom{m}{2}$ is often larger than the number of observations $n$, which imposes a challenge on the model estimation. Thus one common assumption is that both $\mathbf{W}^*$ and $\mathbf{\Omega}^*$ have certain structures. In this paper, without loss of generality, we assume both $\mathbf{W}^*$ and $\mathbf{\Omega}^*$ are sparse, i.e., $\|\mathbf{W}^*\|_{0,0} = s_1^*$ and $\|\mathbf{\Omega}^*\|_{0,0} = s_2^*$. In order to jointly estimate $\mathbf{W}^*$ and $\mathbf{\Omega}^*$, we propose a cardinality constrained maximum likelihood estimator as follows:

$$\min_{\mathbf{W}, \mathbf{\Omega}} -\log|\mathbf{\Omega}| + \frac{1}{n}\operatorname{tr}\big((\mathbf{Y} - \mathbf{X}\mathbf{W})\mathbf{\Omega}(\mathbf{Y} - \mathbf{X}\mathbf{W})^\top\big) \quad \text{subject to} \quad \|\mathbf{W}\|_{0,0} \leq s_1, \|\mathbf{\Omega}\|_{0,0} \leq s_2, \quad (1.3)$$

where $s_1$ and $s_2$ are tuning parameters which control the sparsity of $\mathbf{W}$ and $\mathbf{\Omega}$ respectively. The proposed estimator in (1.3) poses great challenges for both optimization and statistical analysis. The sample loss function in (1.2) is not jointly convex in $\mathbf{W}$ and $\mathbf{\Omega}$. This together with the nonconvex cardinality constraints make our estimator a highly non-convex optimization problem. Moreover, statistical analysis becomes quite challenging for such a non-convex estimator, especially in a finite sample scenario. Many previous studies along this line of research (Kim et al., 2009; Rothman et al., 2010; Lee and Liu, 2012; Wang, 2013) are only able to characterize the asymptotic performance of their estimators. To overcome these challenges, we propose a gradient descent algorithm for solving the nonconvex optimization problem in (1.3). We show that the proposed algorithm is guaranteed to converge to the true parameter matrices $\mathbf{W}^*$ and $\mathbf{\Omega}^*$ at a linear rate up to the statistical error. In particular, the statistical rate of the estimator from our algorithm is $O_P\big(\sqrt{s_1^* \log(dmn)/n} + \sqrt{s_2^* \log(mn)/n}\big)$, which is minimax optimal up to a logarithmic factor. Experiments on both synthetic data and real data verify the superiority of our algorithm and back up our theory.

The remainder of this paper is organized as follows: in Section 2, we briefly review existing work that is relevant to our study. We present the algorithm in Section 3, and the main theory in Section 4. In Section 5, we present the proof of the main theory. In Section 6, we compare the proposed algorithm with existing algorithms on both synthetic data and real-world stock and gene datasets. Finally, we conclude this paper in Section 7.

**Notation.** For a vector $\mathbf{x} \in \mathbb{R}^d$, define vector norm as $\|\mathbf{x}\|_2 = \sqrt{\sum_{i=1}^d x_i^2}$. For a matrix $\mathbf{A} \in \mathbb{R}^{m_1 \times m_2}$, we denote by $\lambda_{\max}(\mathbf{A})$ and $\lambda_{\min}(\mathbf{A})$ the largest and smallest eigenvalue of $\mathbf{A}$, respectively. For a pair of matrices $\mathbf{A}, \mathbf{B}$ with commensurate dimensions, $\langle \mathbf{A}, \mathbf{B} \rangle$ denotes the trace inner product on matrix space that $\langle \mathbf{A}, \mathbf{B} \rangle := \text{trace}(\mathbf{A}^\top \mathbf{B})$. We also define various norms for matrices, including spectral norm $\|\mathbf{A}\|_2 = \max_{\|\mathbf{u}\|_2=1} \|\mathbf{A}\mathbf{u}\|_2$, and the Frobenius norm $\|\mathbf{A}\|_F = \sqrt{\sum_{j=1}^{m_1} \sum_{k=1}^{m_2} A_{jk}^2}$. In addition, we define $\|\mathbf{A}\|_{\infty,\infty} = \max_{1 \leq j \leq m_1, 1 \leq k \leq m_2} |A_{jk}|$, where $A_{jk}$ is the element of $\mathbf{A}$ at row $j$, column $k$. Also for $S \in \{1 \ldots m_1\}$ and $T \in \{1 \ldots m_2\}$, we define $\mathbf{A}_{ST}$ to be the submatrix of $\mathbf{A}$, which is obtained by extracting the appropriate rows and columns in $S$ and $T$ respectively.

## 2 Related Work

A large body of literature (Rothman et al., 2010; Sohn and Kim, 2012; Kim et al., 2009; Lee and Liu, 2012; Wang, 2013) has been devoted to jointly estimating regression coefficients and the sparse precision matrix. In order to solve the resulting nonconvex optimization problem, there exist two families of methods in general. The first family of methods (Rothman et al., 2010; Lee and Liu, 2012) jointly estimates the regression coefficient matrix and precision matrix using alternating optimization algorithms. However, all the theoretical results (Rothman et al., 2010; Lee and Liu, 2012) for alternating minimization algorithms are based on the assumption that there exists a local minimizer that possesses certain good properties, while neither of them is guaranteed to find such a satisfying local minimizer. The second family of methods is identified by their two-step optimization procedure (Cai et al., 2012), in which the regression coefficient matrix is estimated first and the precision matrix estimation is built based on the estimated regression coefficient matrix. However, the two-step approaches cannot fully utilize the interdependency between the regression coefficient matrix and the precision matrix in the estimation process, which would lead to a sub-optimal solution.

In terms of nonconvex optimization technique, Yuan et al. (2013); Jain et al. (2014) proposed and analyzed the gradient descent algorithm with hard thresholding for cardinality constrained optimization problems. However, these algorithms are limited to single optimization variable, and cannot be applied to our problem (contains two optimization variables, i.e., the regression coefficient matrix and the precision matrix). Jain and Tewari (2015) proposed an alternating minimization algorithm for two regression models (i.e., pooled model and seemingly unrelated regression model). However, their regression problems are in the classical regime and are targeted for unconstrained nonconvex optimization problems, which are much less challenging. In addition, alternating minimization has also been analyzed for other models such as matrix factorization (Jain et al., 2013; Arora et al., 2015; Zhao et al., 2015; Zheng and Lafferty, 2015; Chen and Wainwright, 2015; Tu et al., 2015), phase retrieval (Candes et al., 2015) and latent variable models (Balakrishnan et al., 2014; Wang et al., 2015). Yet none of these algorithms and theories can be directly extended to our problem.

## 3 The Proposed Algorithm

In this section, we present a gradient descent based optimization algorithm for solving the proposed estimator in (1.3). The key motivation of the algorithm is that the objective function in (1.3) is convex with respect to $\mathbf{W}$ (resp. $\mathbf{\Omega}$) when the other variable is fixed. We display the algorithm in Algorithm 1. Note that in Algorithm 1, $\nabla_1 f_n$ denotes the gradient of $f_n$ with respect to $\mathbf{W}$, and $\nabla_2 f_n$ denotes its gradient with respect to $\mathbf{\Omega}$.

---
**Algorithm 1** Gradient Descent with Hard Thresholding
---
1: **Input:** Number of iterations $T$, sparsity $s_1, s_2$, step size $\eta_1, \eta_2$, $\mathbf{W}^{\text{init}}$ and $\mathbf{\Omega}^{\text{init}}$
2: **Initial Support Sets:**
   $\widehat{S}_1^{\text{init}} \leftarrow \text{supp}(\mathbf{W}^{\text{init}}, s_1)$, $\widehat{S}_2^{\text{init}} \leftarrow \text{supp}(\mathbf{\Omega}^{\text{init}}, s_2)$
3: **Initial Estimators:**
   $\mathbf{W}^{(0)} \leftarrow \text{HT}(\mathbf{W}^{\text{init}}, \widehat{S}_1^{\text{init}})$, $\mathbf{\Omega}^{(0)} \leftarrow \text{HT}(\mathbf{\Omega}^{\text{init}}, \widehat{S}_2^{\text{init}})$
4: **for** $t = 0$ to $T - 1$ **do**
5:    **Update W:**
   $\mathbf{W}^{(t+0.5)} = \mathbf{W}^{(t)} - \eta_1 \nabla_1 f_n\big(\mathbf{W}^{(t)}, \mathbf{\Omega}^{(t)}\big),$
   $\widehat{S}_1^{(t+0.5)} = \text{supp}(\mathbf{W}^{(t+0.5)}, s_1)$
   $\mathbf{W}^{(t+1)} = \text{HT}(\mathbf{W}^{(t+0.5)}, \widehat{S}_1^{(t+0.5)})$
6:    **Update $\mathbf{\Omega}$:**
   $\mathbf{\Omega}^{(t+0.5)} = \mathbf{\Omega}^{(t)} - \eta_2 \nabla_2 f_n\big(\mathbf{W}^{(t)}, \mathbf{\Omega}^{(t)}\big),$
   $\widehat{S}_2^{(t+0.5)} = \text{supp}(\mathbf{\Omega}^{(t+0.5)}, s_2)$
   $\mathbf{\Omega}^{(t+1)} = \text{HT}(\mathbf{\Omega}^{(t+0.5)}, \widehat{S}_2^{(t+0.5)})$
7: **end for**
8: **Output:** $\widehat{\mathbf{W}} = \mathbf{W}^{(T)}$, $\widehat{\mathbf{\Omega}} = \mathbf{\Omega}^{(T)}$
---

In detail, $\mathbf{W}^{(t+0.5)}$ and $\mathbf{\Omega}^{(t+0.5)}$ are the outputs of gradient descent step. Since $\mathbf{W}^{(t+0.5)}$ and $\mathbf{\Omega}^{(t+0.5)}$ are not necessarily sparse after the gradient descent update, in order to make them sparse, we apply a hard thresholding procedure (Yuan et al., 2013; Jain et al., 2014) right after gradient descent step. The hard thresholding operator is defined as follows:

$$[\text{HT}(\mathbf{W}, S)]_{jk} = \begin{cases} W_{jk}, & \text{if } (j, k) \in S, \\ 0, & \text{if } (j, k) \notin S, \end{cases} \quad (3.1)$$

and $\text{supp}(\mathbf{W}, s)$ is defined as the following index set:

$$\big\{(j, k) : |W_{jk}| \text{ is in the } s \text{ largest elements of all}\big\}. \quad (3.2)$$

The hard thresholding step preserves the entries of $\mathbf{W}^{(t+0.5)}$ and $\mathbf{\Omega}^{(t+0.5)}$ with the top $s_1$ and $s_2$ large magnitudes respectively and sets the rest to zero. This gives rise to $\mathbf{W}^{(t+1)}$ and $\mathbf{\Omega}^{(t+1)}$. Recall that $s_1$ and $s_2$ are tuning parameters that control the sparsity level.

Algorithm 1 provides us an efficient way to solve the non-convex problem using gradient descent with hard thresholding. However, as will be seen in the theoretical analysis, it is guaranteed to convergence to the unknown true parameters only when the initial estimators $\mathbf{W}^{\text{init}}$ and $\mathbf{\Omega}^{\text{init}}$ are sufficiently close to $\mathbf{W}^*$ and $\mathbf{\Omega}^*$ respectively. To reach this requirement, we propose an initialization algorithm in Algorithm 2, which generates initial estimators that are guaranteed to fall in the

---

**Algorithm 2** Initial Estimators

1: **Input:** Regularization parameters $\lambda_1$ and $\lambda_2$
2: **Output:** $\mathbf{W}^{\text{init}} = \text{argmin}_{\mathbf{W}}\{\frac{1}{2n}\|\mathbf{Y} - \mathbf{XW}\|_F^2 + \lambda_1\|\mathbf{W}\|_{1,1}\}$
3: $\qquad \boldsymbol{\Omega}^{\text{init}} = \text{argmin}_{\boldsymbol{\Omega}}\{-\log(|\boldsymbol{\Omega}|) + \text{tr}(\mathbf{S}\boldsymbol{\Omega}) + \lambda_2\|\boldsymbol{\Omega}\|_{1,\text{off}}\}$, where $\mathbf{S} = \frac{1}{n}(\mathbf{Y} - \mathbf{XW}^{\text{init}})^\top(\mathbf{Y} - \mathbf{XW}^{\text{init}})$

---

neighborhood of $\mathbf{W}^*$ and $\boldsymbol{\Omega}^*$ respectively. By combining Algorithm 1 and Algorithm 2, we ensure that our nonconvex optimization algorithm will converge to the true parameters.

## 4 Main Theory

Before we present the main results, we first lay out a series of assumptions, which are essential to establish our theory.

**Assumption 4.1.** The maximum and minimum eigenvalues of $\boldsymbol{\Sigma}^*$ satisfy

$$1/\nu \leq \lambda_{\min}(\boldsymbol{\Sigma}^*) \leq \lambda_{\max}(\boldsymbol{\Sigma}^*) \leq \nu,$$

where $\nu \geq 1$ is an absolute constant. Furthermore we assume that $\nu$ does not increase as $m$ grows to infinity. Since $\boldsymbol{\Omega}^* = \boldsymbol{\Sigma}^{*-1}$, Assumption 4.1 immediately implies that $1/\nu \leq \lambda_{\min}(\boldsymbol{\Omega}^*) \leq \lambda_{\max}(\boldsymbol{\Omega}^*) \leq \nu$. The same assumption has been made in Lee and Liu (2012); Wang (2013); Cai et al. (2012).

**Assumption 4.2.** $\|\mathbf{x}_i\|_2 \leq 1$ for all $i = 1, \ldots, n$. Let $\boldsymbol{\Sigma}_X^* = n^{-1}\mathbb{E}[\mathbf{X}^\top\mathbf{X}]$. There exists a constant $\tau \geq 1$ such that $1/\tau \leq \lambda_{\min}(\boldsymbol{\Sigma}_X^*)$.

Assumption 4.2 states that the minimum eigenvalue of the population covariance matrix of the predictors is bounded away from zero. This assumption is mild and has been widely made in the literature of multivariate regression (Obozinski et al., 2011; Lounici et al., 2009; Negahban and Wainwright, 2011). In fact, since $\|\mathbf{x}_i\|_2 \leq 1$ for all $i = 1, \ldots, n$, it immediately implies that $\lambda_{\max}(\widehat{\boldsymbol{\Sigma}}) \leq 1$.

Now we are going to present our main theorem. To simplify the technical analysis of the proposed algorithm, we focus on its resampling version, which is illustrated in Algorithm 3 in Section 5.

**Theorem 4.3.** Under Assumptions 4.1 and 4.2, let $R = \max\{\|\mathbf{W}^*\|_F, \|\boldsymbol{\Omega}^*\|_F\}$. Let the sparsity parameters be

$$s_1 \geq \max\left\{\frac{100}{9}, \frac{16}{(1/\rho - 1)^2}\right\} \cdot s_1^*, \quad s_2 \geq \max\left\{\frac{100}{9}, \frac{16}{(1/\rho - 1)^2}\right\} \cdot s_2^*, \tag{4.1}$$

Meanwhile, suppose the sample size $n$ is sufficiently large such that

$$\max\{\alpha_1, \alpha_2\} \leq \min\left\{\left(1 - \sqrt{\rho}\right)\frac{R}{4}, \frac{9}{40}R\right\}, \tag{4.2}$$

and

$$n \geq C'' \max\left\{\frac{\nu^2\tau^2 s_1^* \cdot \log(dm)}{R^2}, \frac{\nu^6 s_2^* \cdot \log m}{R^2}, \frac{\nu^4\tau^2 s_1^*\sqrt{s_2^*} \cdot \log(dm)}{R}\right\}, \tag{4.3}$$

Let $\eta_1 \leq 2\nu\tau/(2\nu^2\tau + 1)$, $\eta_2 \leq 1568R^2/(2401R^4 + 256)$. We have with probability at least $1 - 3T/d - 4T^2/n^2$ that

$$\max\left\{\|\mathbf{W}^{(t)} - \mathbf{W}^*\|_F, \|\mathbf{\Omega}^{(t)} - \mathbf{\Omega}^*\|_F\right\} \leq \underbrace{R/4 \cdot \rho^{t/2}}_{\text{Optimization Error}} + \underbrace{\frac{\max\{\alpha_1, \alpha_2\}}{1 - \sqrt{\rho}}}_{\text{Statistical Error}}, \text{ for all } t \in [T] \tag{4.4}$$

where $\rho, \alpha_1$ and $\alpha_2$ are defined as follows

$$\rho = \max\left\{\frac{2\nu^2\tau + R\nu\tau - 1}{2\nu^2\tau + 1}, \frac{2401R^4 + 392R^3 - 1}{2401R^4 + 1}\right\}, \alpha_1 = CR\sqrt{\frac{s_1^* \log(dm)}{n/T}}, \alpha_2 = \frac{C'\nu}{R}\sqrt{\frac{s_2^* \log m}{n/T}}. \tag{4.5}$$

In Theorem 4.3, the result suggests that the estimation error is bounded by two terms: the optimization error term (i.e., the first term on the right hand side of (4.4)), which decays to zero at a linear rate, and the statistical error term (i.e., the second term on the right hand side of (4.4)), which characterizes the the ultimate estimation error in Algorithm 3 when the optimization error term goes to zero as $T$ goes to infinity. In fact, when $T$ is sufficiently large, the optimization error will be smaller than the statistical error, so the total estimator error will be dominated by the statistical error. To be more specific, if we choose the number of iterations $T = C''' \log(n/\log(dm))$ for sufficiently large $C'''$ such that the optimization error term is dominated by the statistical error term, then Theorem 4.3 suggests that the final estimators $\widehat{\mathbf{W}} = \mathbf{W}^{(T)}, \widehat{\mathbf{\Omega}} = \mathbf{\Omega}^{(T)}$ satisfy

$$\max\left\{\|\widehat{\mathbf{W}} - \mathbf{W}^*\|_F, \|\widehat{\mathbf{\Omega}} - \mathbf{\Omega}^*\|_F\right\} \leq \frac{C''''}{1 - \sqrt{\rho}} \max\left\{\sqrt{\frac{s_1^* \log(dm)}{n}}, \nu\sqrt{\frac{s_2^* \log m}{n}}\right\}\sqrt{\log\left(\frac{n}{\log(dm)}\right)}.$$

Comparing with the minimax lower bound for estimating the sparse regression coefficient matrix when the precision matrix $\mathbf{\Omega}^*$ is known: $O_P(\sqrt{s_1^* \log(dm)/n})$ (Obozinski et al., 2011), and the minimax lower bound for estimating the sparse precision matrix when the true regression coefficient matrix $\mathbf{W}^*$ is known: $O_P(\nu\sqrt{s_2^* \log m/n})$ (Obozinski et al., 2011; Cai et al., 2012), there is an additional logarithmic term $\sqrt{\log(n/\log(dm))}$ in our bound. Such a logarithmic factor is introduced by the resampling step in Algorithm 3, since we only utilize $n/T$ samples within each iteration. We expect that it is just an artifact of our proof technique, and such a logarithmic factor can be eliminated by directly analyzing Algorithm 1, which however requires extra technical effort for the analysis. In the following discussion, we ignore the logarithmic factor for simplicity.

**Remark 4.4.** Condition (4.1) in Theorem 4.3 shows that the sparsity parameters $s_1$ and $s_2$ should be chosen to be sufficiently large but meanwhile in the same order as the true sparsity level $s_1^*$ and $s_2^*$ respectively. This ensures that the extra error caused by hard thresholding step can be upper bounded. In addition, in order to make sure the contraction parameter of the linear convergence $\rho < 1$, we require $R$ to be sufficiently small. This can be achieved by scaling the data easily. Also, the condition on the sample size $n$ in (4.3) is due to the initialization procedure in Algorithm 2. When $n$ satisfies (4.3), the initial estimators generated by Algorithm 2 will lie in the basin of contraction for the true parameter matrices $\mathbf{W}^*$ and $\mathbf{\Omega}^*$ respectively. This ensures the linear rate of convergence of Algorithm 1, as stated in the above theorem.

**Remark 4.5.** Comparing with the rates achieved by the initialization estimators in Algorithm 2, Algorithm 1 clearly improves the estimation result for both $\mathbf{W}^*$ and $\mathbf{\Omega}^*$. Specifically, for estimating $\mathbf{\Omega}^*$, the initialization algorithm can only obtain a convergence rate of $O_P(\nu^3\sqrt{s_2^*\log m/n} + \nu^4\tau^2 s_1^*\sqrt{s_2^*} \cdot \log(dm)/n)$ (See Lemma 5.10), while Algorithm 1 achieves a sharper rate of $O_P(\nu\sqrt{s_2^*\log m/n})$. For estimating $\mathbf{W}^*$, one may think that Algorithm 1 and the initialization algorithm achieve the same rate of $O_P(\sqrt{s_1^*\log(dm)/n})$. However, this is not true because we should also consider the problem-dependent multiplicative factors $\tau$ and $\nu$ ($\tau > 1$ and $\nu > 1$). To be precise, the rate of the initial estimator $\mathbf{W}^{\text{init}}$ is $O_P(\tau\nu\sqrt{s_1^*\log(dm)/n})$ (See Lemma 5.9), while the final estimator returned by Algorithm 1 achieves a faster rate of $O_P(\sqrt{s_1^*\log(dm)/n})$, which is independent of $\tau$ and $\nu$. Note that $\tau$ and $\nu$ can be very large in general, hence the estimators returned by Algorithm 1 improve the initial estimators substantially.

## 5 Proof of the Main Theory

In this section, we will provide a detailed proof for the main theory. To simplify the technical analysis of the proposed algorithm, we focus on its resampling version, which is illustrated in Algorithm 3. The key idea is to split the whole dataset into $T$ pieces and use a fresh piece of data in each iteration of Algorithm 3.

---
**Algorithm 3** Gradient Descent with Hard Thresholding (Resampling Version)
---
1: **Parameters:** Maximal number of iterations $T$, Sparsity parameters $s_1, s_2$, Step size parameters $\eta_1, \eta_2$
2: **Initial Support Sets:** $\widehat{S}_1^{\text{init}} \leftarrow \text{supp}(\mathbf{W}^{\text{init}}, s_1)$, $\widehat{S}_2^{\text{init}} \leftarrow \text{supp}(\mathbf{\Omega}^{\text{init}}, s_2)$
Split the Dataset into $T$ Subsets of Size $n/T$ (Assuming $n/T$ is an integer)
3: **Initial Estimators:**
$\mathbf{W}^{(0)} \leftarrow \text{HT}(\mathbf{W}^{\text{init}}, \widehat{S}_1^{\text{init}})$, $\mathbf{\Omega}^{(0)} \leftarrow \text{HT}(\mathbf{\Omega}^{\text{init}}, \widehat{S}_2^{\text{init}})$
4: **for** $t = 0$ to $T-1$ **do**
5:    **Update W:**
$\mathbf{W}^{(t+0.5)} = \mathbf{W}^{(t)} - \eta_1 \nabla_1 f_{n/T}(\mathbf{W}^{(t)}, \mathbf{\Omega}^{(t)})$ with the $t$-th Data Subset
$\widehat{S}_1^{(t+0.5)} = \text{supp}(\mathbf{W}^{(t+0.5)}, s_1)$
$\mathbf{W}^{(t+1)} = \text{HT}(\mathbf{W}^{(t+0.5)}, \widehat{S}_1^{(t+0.5)})$
6:    **Update $\mathbf{\Omega}$:**
$\mathbf{\Omega}^{(t+0.5)} = \mathbf{\Omega}^{(t)} - \eta_2 \nabla_2 f_{n/T}(\mathbf{W}^{(t)}, \mathbf{\Omega}^{(t)})$ with the $t$-th Data Subset
$\widehat{S}_2^{(t+0.5)} = \text{supp}(\mathbf{\Omega}^{(t+0.5)}, s_2)$
$\mathbf{\Omega}^{(t+1)} = \text{HT}(\mathbf{\Omega}^{(t+0.5)}, \widehat{S}_2^{(t+0.5)})$
7: **end for**
8: **Output:** $\widehat{\mathbf{W}} = \mathbf{W}^{(T)}$, $\widehat{\mathbf{\Omega}} = \mathbf{\Omega}^{(T)}$
---

Before we begin our proof, we first define $\mathbb{B}_F(\mathbf{W}^*; r) = \{\mathbf{W} \in \mathbb{R}^{d \times m} : \|\mathbf{W} - \mathbf{W}^*\|_F \le r\}$. Similarly we define $\mathbb{B}_F(\mathbf{\Omega}^*; r) = \{\mathbf{\Omega} \in \mathbb{R}^{m \times m} : \|\mathbf{\Omega} - \mathbf{\Omega}^*\|_F \le r\}$. Moreover we assume that $R = \max\{\|\mathbf{W}^*\|_F, \|\mathbf{\Omega}^*\|_F\} = 4r$ in the following proof. Now we introduce several lemmas which are essential to the proof of our main theorem.

**Lemma 5.1.** Under Assumptions 4.1 and 4.2, for any $\mathbf{W}', \mathbf{W} \in \mathbb{B}_F(\mathbf{W}^*; r)$, the population loss function $f(\cdot, \mathbf{\Omega}^*)$ is $1/(\nu\tau)$-strongly convex and $(2\nu)$-smooth, i.e.,

$$\frac{1}{2\nu\tau}\|\mathbf{W}' - \mathbf{W}\|_F^2 \leq f(\mathbf{W}', \mathbf{\Omega}^*) - f(\mathbf{W}, \mathbf{\Omega}^*) - \nabla_1 f(\mathbf{W}, \mathbf{\Omega}^*)^\top (\mathbf{W}' - \mathbf{W}) \leq \nu \|\mathbf{W}' - \mathbf{W}\|_F^2.$$

**Lemma 5.2.** For any $\mathbf{\Omega}', \mathbf{\Omega} \in \mathbb{B}_F(\mathbf{\Omega}^*; r)$, the population loss function $f(\mathbf{W}^*, \cdot)$ is $[\|\mathbf{\Omega}^*\|_F + 3r]^{-2}$-strongly convex and $[\|\mathbf{\Omega}^*\|_F + 3r]^2$-smooth, i.e.,

$$\frac{\|\mathbf{\Omega}' - \mathbf{\Omega}\|_F^2}{2[\|\mathbf{\Omega}^*\|_F + 3r]^2} \leq f(\mathbf{W}^*, \mathbf{\Omega}') - f(\mathbf{W}^*, \mathbf{\Omega}) - \nabla_2 f(\mathbf{W}^*, \mathbf{\Omega})^\top(\mathbf{\Omega}' - \mathbf{\Omega}) \leq \frac{[\|\mathbf{\Omega}^*\|_F + 3r]^2}{2}\|\mathbf{\Omega}' - \mathbf{\Omega}\|_F^2.$$

Lemmas 5.1 and 5.2 indicate that when one of variables is fixed as true variable (i.e., $\mathbf{W}^*$ or $\mathbf{\Omega}^*$), the population function $f$ is both strongly convex and smooth with respect to the other variable. These conclusions ensure that the standard convex optimization results for strongly convex and smooth objective functions (Nesterov, 2004) can be applied to function $f$ as long as one of the variables takes its true value.

**Lemma 5.3.** Suppose Assumptions 4.1 and 4.2 hold. For the true parameter $\mathbf{\Omega}^*$ and any $\mathbf{\Omega} \in \mathbb{B}_F(\mathbf{\Omega}^*; r)$, the gradient difference $\nabla_1 f(\mathbf{W}, \mathbf{\Omega}^*) - \nabla_1 f(\mathbf{W}, \mathbf{\Omega})$ satisfies

$$\|\nabla_1 f(\mathbf{W}, \mathbf{\Omega}^*) - \nabla_1 f(\mathbf{W}, \mathbf{\Omega})\|_F \leq 2r \cdot \|\mathbf{\Omega}^* - \mathbf{\Omega}\|_F. \tag{5.1}$$

For true parameter $\mathbf{W}^*$ and any $\mathbf{W} \in \mathbb{B}_F(\mathbf{W}^*; r)$, the gradient difference $\nabla_2 f(\mathbf{W}^*, \mathbf{\Omega}) - \nabla_2 f(\mathbf{W}, \mathbf{\Omega})$ satisfies

$$\|\nabla_2 f(\mathbf{W}^*, \mathbf{\Omega}) - \nabla_2 f(\mathbf{W}, \mathbf{\Omega})\|_F \leq r \cdot \|\mathbf{W}^* - \mathbf{W}\|_F. \tag{5.2}$$

Lemma 5.3 suggests the gradients satisfy Lipschitz property with respect to $\mathbf{\Omega}$ and $\mathbf{W}$. Note that this Lipschitz property only holds between the true parameter ($\mathbf{W}^*$ or $\mathbf{\Omega}^*$) and arbitrary parameter in the neighborhood of the true parameter ($\mathbf{W} \in \mathbb{B}_F(\mathbf{W}^*; r)$ or $\mathbf{\Omega} \in \mathbb{B}_F(\mathbf{\Omega}^*; r)$). Given Lemma 5.3, standard convex optimization results can be adapted to analyze $f(\cdot, \mathbf{\Omega})$ for any $\mathbf{\Omega} \in \mathbb{B}_F(\mathbf{\Omega}^*; r)$ and $f(\mathbf{W}, \cdot)$ for any $\mathbf{W} \in \mathbb{B}_F(\mathbf{W}^*; r)$.

**Lemma 5.4.** For any fixed $\mathbf{W} \in \mathbb{B}_F(\mathbf{W}^*; r)$ and $\mathbf{\Omega} \in \mathbb{B}_F(\mathbf{\Omega}^*; r)$, with probability at least $1 - 2T/d$, we have

$$\|\nabla_1 f(\mathbf{W}, \mathbf{\Omega}) - \nabla_1 f_{n/T}(\mathbf{W}, \mathbf{\Omega})\|_{\infty,\infty} \leq \epsilon_1, \tag{5.3}$$

where $\epsilon_1 = C(r + \|\mathbf{\Omega}^*\|_F)\nu\sqrt{\log(dm) \cdot T/n}$. And with probability at least $1 - 2T/d - 4T^2/n^2$, we have

$$\|\nabla_2 f(\mathbf{W}, \mathbf{\Omega}) - \nabla_2 f_{n/T}(\mathbf{W}, \mathbf{\Omega})\|_{\infty,\infty} \leq \epsilon_2, \tag{5.4}$$

where $\epsilon_2 = C'r\nu\sqrt{(\log m) \cdot T/n}$.

Lemma 5.4 characterizes the difference between the gradient of the population loss function and the gradient of the sample loss function, in terms of $\ell_{\infty,\infty}$ norm. We use $\ell_{\infty,\infty}$ norm for it is the dual norm of $\ell_{1,1}$ norm and characterizes a refined entry-wise error with a sharp convergence rate.

To further continue our proof procedure, here we define the gradient descent update process for the sample loss function as follows:

$$M_{1,n/T}(\mathbf{W}, \mathbf{\Omega}) = \mathbf{W} - \eta_1 \nabla_1 f_{n/T}(\mathbf{W}, \mathbf{\Omega}), \ M_{2,n/T}(\mathbf{W}, \mathbf{\Omega}) = \mathbf{\Omega} - \eta_2 \nabla_2 f_{n/T}(\mathbf{W}, \mathbf{\Omega}).$$

Also we defined $M_1(\mathbf{W}, \mathbf{\Omega})$, $M_2(\mathbf{W}, \mathbf{\Omega})$ as the results of gradient descent update for the population loss function with respect to $\mathbf{W}$ and $\mathbf{\Omega}$ as follows:

$$M_1(\mathbf{W}, \mathbf{\Omega}) = \mathbf{W} - \eta_1 \nabla_1 f(\mathbf{W}, \mathbf{\Omega}), \ M_2(\mathbf{W}, \mathbf{\Omega}) = \mathbf{\Omega} - \eta_2 \nabla_2 f(\mathbf{W}, \mathbf{\Omega}),$$

where $f(\mathbf{W}, \mathbf{\Omega}) = \mathbb{E}[f_n(\mathbf{W}, \mathbf{\Omega})]$ is the population loss function. Recall that the $\mathrm{HT}(\cdot, \cdot)$ and $\mathrm{supp}(\cdot, \cdot)$ defined in (3.1) and (3.2), we further define:

$$\overline{\mathbf{W}}^{(t+0.5)} = M_1(\mathbf{W}^{(t)}, \mathbf{\Omega}^{(t)}), \ \overline{\mathbf{W}}^{(t+1)} = \mathrm{HT}(\overline{\mathbf{W}}^{(t+0.5)}, \widehat{S}_1^{(t+0.5)}), \quad (5.5)$$
$$\overline{\mathbf{\Omega}}^{(t+0.5)} = M_2(\mathbf{W}^{(t)}, \mathbf{\Omega}^{(t)}), \ \overline{\mathbf{\Omega}}^{(t+1)} = \mathrm{HT}(\overline{\mathbf{\Omega}}^{(t+0.5)}, \widehat{S}_2^{(t+0.5)}).$$

Our subsequent proof will try to characterize the upper bound for $\|\overline{\mathbf{W}}^{(t+1)} - \mathbf{W}^*\|_F$ and accordingly $\|\overline{\mathbf{\Omega}}^{(t+1)} - \mathbf{\Omega}^*\|_F$, in two steps. First we characterize the relationship between $\|\overline{\mathbf{W}}^{(t+0.5)} - \mathbf{W}^*\|_F$ and $\|\overline{\mathbf{W}}^{(t)} - \mathbf{W}^*\|_F$ (resp. $\|\overline{\mathbf{\Omega}}^{(t+0.5)} - \mathbf{\Omega}^*\|_F$ and $\|\overline{\mathbf{\Omega}}^{(t)} - \mathbf{\Omega}^*\|_F$). Then we will describe the relationship between $\|\overline{\mathbf{W}}^{(t+1)} - \mathbf{W}^*\|_F$ and $\|\overline{\mathbf{W}}^{(t+0.5)} - \mathbf{W}^*\|_F$ (resp. $\|\overline{\mathbf{\Omega}}^{(t+1)} - \mathbf{\Omega}^*\|_F$ and $\|\overline{\mathbf{\Omega}}^{(t+0.5)} - \mathbf{\Omega}^*\|_F$). The next two lemmas will accomplish the first step.

**Lemma 5.5.** Under Assumptions 4.1 and 4.2, suppose that $\mathbf{W} \in \mathbb{B}_F(\mathbf{W}^*; r)$, $\mathbf{\Omega} \in \mathbb{B}_F(\mathbf{\Omega}^*; r)$, then Algorithm 1 with step sizes $\eta_1 \leq 2\nu\tau/(2\nu^2\tau + 1)$, $\eta_2 \leq 1568R^2/(2401R^4 + 256)$ satisfies

$$\|\overline{\mathbf{W}}^{(t+0.5)} - \mathbf{W}^*\|_F \leq \frac{2\nu^2\tau - 1}{2\nu^2\tau + 1} \cdot \|\mathbf{W}^{(t)} - \mathbf{W}^*\|_F + \frac{R\nu\tau}{2\nu^2\tau + 1} \cdot \|\mathbf{\Omega}^{(t)} - \mathbf{\Omega}^*\|_F.$$

Similarly, we have the following lemma, which establishes the contraction of $\|\overline{\mathbf{\Omega}}^{(t+0.5)} - \mathbf{\Omega}^*\|_F$ with respect to $\|\mathbf{\Omega}^{(t)} - \mathbf{\Omega}^*\|_F$ and $\|\mathbf{W}^{(t)} - \mathbf{W}^*\|_F$.

**Lemma 5.6.** Suppose that $\mathbf{W} \in \mathbb{B}_F(\mathbf{W}^*; r)$, $\mathbf{\Omega} \in \mathbb{B}_F(\mathbf{\Omega}^*; r)$, then Algorithm 1 with step size $\eta_1 \leq 2\nu\tau/(2\nu^2\tau + 1)$, $\eta_2 \leq 1568R^2/(2401R^4 + 256)$ satisfies

$$\|\overline{\mathbf{\Omega}}^{(t+0.5)} - \mathbf{\Omega}^*\|_F \leq \frac{2401R^4 - 1}{2401R^4 + 1} \cdot \|\mathbf{\Omega}^{(t)} - \mathbf{\Omega}^*\|_F + \frac{392R^3}{2401R^4 + 1} \cdot \|\mathbf{W}^{(t)} - \mathbf{W}^*\|_F.$$

Lemmas 5.5 and 5.6 are crucial to the proof of Theorem 4.3. They establish the bound for the estimation error by conducting one-step gradient descent on the population loss function. In addition, the connections we need for proving the main theorem by induction are successfully built by these lemmas, through the estimation error of each parameter before the gradient descent.

Next we need the following lemma to characterize the relationship between the estimation error by conducting one-step gradient descent on the population loss function after hard thresholding (i.e., $\|\overline{\mathbf{W}}^{(t+1)} - \mathbf{W}^*\|_F$) and before hard thresholding (i.e., $\|\overline{\mathbf{W}}^{(t+0.5)} - \mathbf{W}^*\|_F$).

**Lemma 5.7** (Lemma 5.1 in Wang et al. (2015)). Suppose that we have

$$\|\overline{\mathbf{W}}^{(t+0.5)} - \mathbf{W}^*\|_F \leq \kappa \cdot \|\mathbf{W}^*\|_F,$$

for some $\kappa \in (0, 1)$, and

$$s_1 \geq \frac{4 \cdot (1+\kappa)^2}{(1-\kappa)^2} \cdot s_1^*, \tag{5.6}$$

$$\sqrt{s_1} \cdot \|\mathbf{W}^{(t+0.5)} - \overline{\mathbf{W}}^{(t+0.5)}\|_{\infty,\infty} \leq \frac{(1-\kappa)^2}{2(1+\kappa)} \cdot \|\mathbf{W}^*\|_F, \tag{5.7}$$

then it holds that

$$\|\overline{\mathbf{W}}^{(t+1)} - \mathbf{W}^*\|_F \leq \frac{C \cdot \sqrt{s_1^*}}{\sqrt{1-\kappa}} \cdot \|\mathbf{W}^{(t+0.5)} - \overline{\mathbf{W}}^{(t+0.5)}\|_{\infty,\infty}$$
$$+ \left(1 + 4\sqrt{s_1^*/s_1}\right)^{1/2} \cdot \|\overline{\mathbf{W}}^{(t+0.5)} - \mathbf{W}^*\|_F. \tag{5.8}$$

Lemma 5.7 is vital to the proof of Theorem 4.3. In detail, the assumption in (5.6) ensures the sparsity parameter $s_1$ is set to be sufficiently large, and since $\|\mathbf{W}^{(t+0.5)} - \overline{\mathbf{W}}^{(t+0.5)}\|_{\infty,\infty}$ (See the proof of Lemma 5.8) is proportional to the statistical error, the assumption in (5.7) makes sure that the statistical error term is sufficiently small. Lemma 5.7 suggests that, after hard thresholding step, the estimation error by conducting one step of gradient update for the population loss function can still be controlled by the combination of the statistical error and the estimation error before hard thresholding. Intuitively speaking, the hard thresholding step imposes additional error to the estimation error, but meanwhile, it can be controlled. In particular, such additional error effect reduces as the sparsity parameter $s_1$ increases. In other words, the larger $s_1$ is, the fewer relevant entries are wrongly eliminated by the hard thresholding step, hence the smaller additional error will be introduced. Similar lemma holds for $\|\overline{\mathbf{\Omega}}^{(t+1)} - \mathbf{\Omega}^*\|_F$ and $\|\overline{\mathbf{\Omega}}^{(t+0.5)} - \mathbf{\Omega}^*\|_F$.

**Lemma 5.8.** Under Assumptions 4.1 and 4.2, for any $\mathbf{W} \in \mathbb{B}_F(\mathbf{W}^*; r)$, following the update process in Algorithm 1, we have

$$\left\|\mathrm{HT}(\mathbf{W}^{(t+0.5)}, \widehat{S}_1^{(t+0.5)}) - \mathrm{HT}(\overline{\mathbf{W}}^{(t+0.5)}, \widehat{S}_1^{(t+0.5)})\right\|_F \leq \frac{2\nu\tau}{2\nu^2\tau + 1}\sqrt{s_1} \cdot \epsilon_1. \tag{5.9}$$

Similarly, for any $\mathbf{\Omega} \in \mathbb{B}_F(\mathbf{\Omega}^*; r)$, following the update process in Algorithm 1, we have

$$\left\|\mathrm{HT}(\mathbf{\Omega}^{(t+0.5)}, \widehat{S}_2^{(t+0.5)}) - \mathrm{HT}(\overline{\mathbf{\Omega}}^{(t+0.5)}, \widehat{S}_2^{(t+0.5)})\right\|_F \leq \frac{1568R^2}{2401R^4 + 256}\sqrt{s_2} \cdot \epsilon_2. \tag{5.10}$$

Here $\epsilon_1$ and $\epsilon_2$ are defined in Lemma 5.4.

Lemma 5.8 further ensures that the difference between the gradient update for the sample loss function after hard thresholding and the gradient update for the population loss function after hard thresholding can be bounded by statistical error with some constant scalers.

**Lemma 5.9.** Under Assumption 4.2, if the regularization parameter $\lambda_1$ in Algorithm 2 satisfies $\lambda_1 \geq C\nu\sqrt{\log(dm)/n}$, then with probability at least $1 - 1/(dm) - 1/n^2$, it holds that

$$\|\mathbf{W}^{\mathrm{init}} - \mathbf{W}^*\|_F \leq C'\nu\tau\sqrt{s_1^* \cdot \log(dm)/n}. \tag{5.11}$$

**Lemma 5.10.** Under Assumptions 4.1 and 4.2, if the regularization parameter $\lambda_2$ in Algorithm 2 satisfies $\lambda_2 \geq C\nu\sqrt{\log m/n} + C'\nu^2\tau^2 s_1^* \cdot (\log m + \log d)/n$, then with probability at least $1 - 1/d - 4/n^2$, it holds that

$$\|\mathbf{\Omega}^{\text{init}} - \mathbf{\Omega}^*\|_F \leq C'\nu^3\sqrt{\frac{s_2^* \cdot \log m}{n}} + C''\nu^4\tau^2 s_1^* \sqrt{s_2^*} \cdot \frac{\log m + \log d}{n}.$$

Lemmas 5.9 and 5.10 provide us the statistical guarantees for our initial estimators in Algorithm 2. Specifically, the initialization algorithm achieves a convergence rate of $O_P\big(\nu\tau\sqrt{s_1^* \cdot \log(dm)/n}\big)$ for estimating $\mathbf{W}^*$ and a convergence rate of $O_P\big(\nu^3\sqrt{s_2^*\log m/n} + \nu^4\tau^2 s_1^*\sqrt{s_2^*} \cdot \log(dm)/n\big)$. By comparing these rates to the minimax lower bound for estimating the sparse regression coefficient matrix when the precision matrix $\mathbf{\Omega}^*$ is known (Obozinski et al., 2011), and the minimax lower bound for estimating the sparse precision matrix when the true regression coefficient matrix $\mathbf{W}^*$ is known (Obozinski et al., 2011; Cai et al., 2012), we can see that neither of the rates for the initial estimators is minimax optimal due to the extra multiplicative factors.

Now we have gathered everything we need and we are ready to present the proof of the main theorem.

*Proof of Theorem 4.3.* We will prove the main theorem by mathematical induction, which contains two steps. In step one we will prove that $\|\mathbf{W}^{(0)} - \mathbf{W}^*\|_F \leq r$ and $\|\mathbf{\Omega}^{(0)} - \mathbf{\Omega}^*\|_F \leq r$, where $r = R/4$. And in step two we try to expand our conclusion to any $t$ given the information about $(t-1)$-th iteration.

**Step 1:** According to Lemma 5.9, we have that

$$\|\mathbf{W}^{\text{init}} - \mathbf{W}^*\|_F \leq C_m \nu\tau \cdot \sqrt{s_1^* \cdot \log(dm)/n}.$$

Thus according to the theorem condition (4.3), we get

$$\|\mathbf{W}^{\text{init}} - \mathbf{W}^*\|_F \leq R/8.$$

The same argument applies to the initial estimator $\mathbf{\Omega}^{\text{init}}$. According to Lemma 5.10, we have that

$$\|\mathbf{\Omega}^{\text{init}} - \mathbf{\Omega}^*\|_F \leq C_{g_1}\nu^3 \cdot \sqrt{s_2^* \cdot \log m/n} + C_{g_2}\nu^4\tau^2 s_1^*\sqrt{s_2^*} \cdot \log(dm)/n.$$

Therefore, according to theorem condition (4.3) we have

$$\|\mathbf{\Omega}^{\text{init}} - \mathbf{\Omega}^*\|_F \leq R/8.$$

Recall that $r = R/4$, we have

$$\|\mathbf{W}^{\text{init}} - \mathbf{W}^*\|_F \leq r/2, \ \|\mathbf{\Omega}^{\text{init}} - \mathbf{\Omega}^*\|_F \leq r/2.$$

Since $\mathbf{W}^{(0)}$ is the result of $\mathbf{W}^{\text{init}}$ after hard thresholding, by Lemma 5.7, we have

$$\|\mathbf{W}^{(0)} - \mathbf{W}^*\|_F \leq \big(1 + 4\sqrt{s_1^*/s_1}\big)^{1/2} \cdot \|\mathbf{W}^{\text{init}} - \mathbf{\Omega}^*\|_F \leq \big(1 + 4\sqrt{s_1^*/s_1}\big)^{1/2} \cdot \frac{r}{2}, \qquad (5.12)$$

where the first inequality holds due to that the first item in (5.8) of Lemma 5.7 does not exists since there is no gradient descent update yet. Note that from theorem condition (4.1) we have, $\sqrt{s_1^*/s_1} \leq 9/100 \leq 1/4$, thus (5.12) further implies that

$$\|\mathbf{W}^{(0)} - \mathbf{W}^*\|_F \leq \left(1 + 4\sqrt{1/4}\right)^{1/2} \cdot \frac{r}{2} \leq r.$$

Similarly, we can prove that for $\mathbf{\Omega}$, we have $\|\mathbf{\Omega}^{(0)} - \mathbf{\Omega}^*\|_F \leq r$.

**Step 2:** Suppose that $\mathbf{W}^{(t-1)} \in \mathbb{B}_F(\mathbf{W}^*; r)$, $\mathbf{\Omega}^{(t-1)} \in \mathbb{B}_F(\mathbf{\Omega}^*; r)$. Consider the estimation error of $t$-th iteration:

$$\begin{aligned}
\|\mathbf{W}^{(t+1)} - \mathbf{W}^*\|_F &= \|\mathrm{HT}(\mathbf{W}^{(t+0.5)}, \widehat{S}_1^{(t+0.5)}) - \mathbf{W}^*\| \\
&\leq \|\mathrm{HT}(\mathbf{W}^{(t+0.5)}, \widehat{S}_1^{(t+0.5)}) - \mathrm{HT}(\overline{\mathbf{W}}^{(t+0.5)}, \widehat{S}_1^{(t+0.5)})\|_F + \|\mathrm{HT}(\overline{\mathbf{W}}^{(t+0.5)}, \widehat{S}_1^{(t+0.5)}) - \mathbf{W}^*\|_F \\
&= \underbrace{\|\mathrm{HT}(\mathbf{W}^{(t+0.5)}, \widehat{S}_1^{(t+0.5)}) - \mathrm{HT}(\overline{\mathbf{W}}^{(t+0.5)}, \widehat{S}_1^{(t+0.5)})\|_F}_{I_1} + \underbrace{\|\overline{\mathbf{W}}^{(t+1)} - \mathbf{W}^*\|_F}_{I_2},
\end{aligned} \quad (5.13)$$

where the inequality holds due to triangle inequality. For the two terms in estimation error $I_1$ and $I_2$, the first item is bounded by Lemma 5.8, where,

$$\left\|\mathrm{HT}(\mathbf{W}^{(t+0.5)}, \widehat{S}_1^{(t+0.5)}) - \mathrm{HT}(\overline{\mathbf{W}}^{(t+0.5)}, \widehat{S}_1^{(t+0.5)})\right\|_F \leq \frac{2\nu\tau}{2\nu^2\tau + 1}\sqrt{s_1} \cdot \epsilon_1. \quad (5.14)$$

The second term $I_2$ can be bounded by combining Lemmas 5.5 and 5.7. Submitting the conclusion from Lemma 5.5 into Lemma 5.7 with $\kappa = 1/4$ we have

$$\begin{aligned}
\|\overline{\mathbf{W}}^{(t+1)} - \mathbf{W}^*\|_F &\leq C'' \cdot \sqrt{s_1^*} \cdot \|\mathbf{W}^{(t+0.5)} - \overline{\mathbf{W}}^{(t+0.5)}\|_{\infty,\infty} \\
&\quad + \left(1 + 4\sqrt{s_1^*/s_1}\right)^{1/2} \cdot \left[\frac{2\nu^2\tau - 1}{2\nu^2\tau + 1} \cdot \|\mathbf{W}^{(t)} - \mathbf{W}^*\|_F + \frac{R\nu\tau}{2\nu^2\tau + 1} \cdot \|\mathbf{\Omega}^{(t)} - \mathbf{\Omega}^*\|_F\right].
\end{aligned} \quad (5.15)$$

By adding up (5.14) with (5.15), we combine term $I_1$ and $I_2$ together and get

$$\begin{aligned}
\|\mathbf{W}^{(t+1)} - \mathbf{W}^*\|_F &= \frac{2\nu\tau}{2\nu^2\tau + 1}\sqrt{s_1} \cdot \epsilon_1 + C'' \cdot \sqrt{s_1^*} \cdot \|\mathbf{W}^{(t+0.5)} - \overline{\mathbf{W}}^{(t+0.5)}\|_{\infty,\infty} \\
&\quad + \left(1 + 4\sqrt{s_1^*/s_1}\right)^{1/2} \cdot \left[\frac{2\nu^2\tau - 1}{2\nu^2\tau + 1} \cdot \|\mathbf{W}^{(t)} - \mathbf{W}^*\|_F + \frac{R\nu\tau}{2\nu^2\tau + 1} \cdot \|\mathbf{\Omega}^{(t)} - \mathbf{\Omega}^*\|_F\right] \\
&\leq \frac{2\nu\tau}{2\nu^2\tau + 1}\left(\sqrt{s_1} + C'' \cdot \sqrt{s_1^*}\right) \cdot \epsilon_1 \\
&\quad + \left(1 + 4\sqrt{s_1^*/s_1}\right)^{1/2} \cdot \left[\frac{2\nu^2\tau - 1}{2\nu^2\tau + 1} \cdot \|\mathbf{W}^{(t)} - \mathbf{W}^*\|_F + \frac{R\nu\tau}{2\nu^2\tau + 1} \cdot \|\mathbf{\Omega}^{(t)} - \mathbf{\Omega}^*\|_F\right],
\end{aligned}$$

where the inequality follows from similar proving procedure as in Lemma 5.8. Similarly, we establish the bound for $\|\mathbf{\Omega}^{(t+1)} - \mathbf{\Omega}^*\|_F$ following the conclusions in Lemmas 5.6, 5.7 and 5.8 accordingly,

$$\begin{aligned}
\|\mathbf{\Omega}^{(t+1)} - \mathbf{\Omega}^*\|_F &\leq \frac{1568R^2}{2401R^4 + 256}\left(\sqrt{s_2} + C'' \cdot \sqrt{s_2^*}\right) \cdot \epsilon_2 \\
&\quad + \left(1 + 4\sqrt{s_2^*/s_2}\right)^{1/2} \cdot \left[\frac{2401R^4 - 1}{2401R^4 + 1} \cdot \|\mathbf{\Omega}^{(t)} - \mathbf{\Omega}^*\|_F + \frac{392R^3}{2401R^4 + 1} \cdot \|\mathbf{W}^{(t)} - \mathbf{W}^*\|_F\right].
\end{aligned}$$

Notice that we prove the resampling version of the theorem, thus by incorporating Lemma 5.4, we have

$$\|\mathbf{W}^{(t+1)} - \mathbf{W}^*\|_F \leq CR\big(\sqrt{s_1} + C''\sqrt{s_1^*}\big) \cdot \sqrt{\log(dm) \cdot T/n}$$
$$+ \Big(1 + 4\sqrt{s_1^*/s_1}\Big)^{1/2} \cdot \left[\frac{2\nu^2\tau - 1}{2\nu^2\tau + 1} \cdot \|\mathbf{W}^{(t)} - \mathbf{W}^*\|_F + \frac{R\nu\tau}{2\nu^2\tau + 1} \cdot \|\mathbf{\Omega}^{(t)} - \mathbf{\Omega}^*\|_F\right], \quad (5.16)$$

and

$$\|\mathbf{\Omega}^{(t+1)} - \mathbf{\Omega}^*\|_F \leq \frac{C'\nu}{R}\big(\sqrt{s_2} + C'' \cdot \sqrt{s_2^*}\big) \cdot \sqrt{\log m \cdot T/n}$$
$$+ \Big(1 + 4\sqrt{s_2^*/s_2}\Big)^{1/2} \cdot \left[\frac{2401R^4 - 1}{2401R^4 + 1} \cdot \|\mathbf{\Omega}^{(t)} - \mathbf{\Omega}^*\|_F + \frac{392R^3}{2401R^4 + 1} \cdot \|\mathbf{W}^{(t)} - \mathbf{W}^*\|_F\right]. \quad (5.17)$$

Note that by our assumptions $s_1 \geq 16 \cdot (1/\rho - 1)^{-2} \cdot s_1^*$ and $s_2 \geq 16 \cdot (1/\rho - 1)^{-2} \cdot s_2^*$, we have

$$\max\left\{\Big(1 + 4 \cdot \sqrt{s_1^*/s_1}\Big)^{1/2}, \Big(1 + 4 \cdot \sqrt{s_2^*/s_2}\Big)^{1/2}\right\} \leq 1/\sqrt{\rho}.$$

Thus by combining (5.16) together with (5.17), we get

$$\max\big\{\|\mathbf{W}^{(t+1)} - \mathbf{W}^*\|_F, \|\mathbf{\Omega}^{(t+1)} - \mathbf{\Omega}^*\|_F\big\} \leq \sqrt{\rho} \cdot \max\big\{\|\mathbf{W}^{(t)} - \mathbf{W}^*\|_F, \|\mathbf{\Omega}^{(t)} - \mathbf{\Omega}^*\|_F\big\} + \max\{\alpha_1, \alpha_2\}, \quad (5.18)$$

where $\rho, \alpha_1$ and $\alpha_2$ are defined in (4.5).

For simplicity, if we denote $M^{(t)} = \max\big\{\|\mathbf{W}^{(t)} - \mathbf{W}^*\|_F, \|\mathbf{\Omega}^{(t)} - \mathbf{\Omega}^*\|_F\big\}$, $\zeta = \max\{\alpha_1, \alpha_2\}$ and take one step back from iteration $t+1$ to $t$, then (5.18) can be rewritten as:

$$M^{(t)} \leq \sqrt{\rho} \cdot M^{(t-1)} + \zeta. \quad (5.19)$$

Since we have $\mathbf{W}^{(t-1)} \in \mathbb{B}_F(\mathbf{W}^*; r)$, $\mathbf{\Omega}^{(t-1)} \in \mathbb{B}_F(\mathbf{\Omega}^*; r)$, by (5.19), it immediately implies that

$$M^{(t-1)} = \max\big\{\|\mathbf{W}^{(t-1)} - \mathbf{W}^*\|_F, \|\mathbf{\Omega}^{(t-1)} - \mathbf{\Omega}^*\|_F\big\} \leq r.$$

Given the theorem condition (4.2), we can easily derive that

$$\zeta \leq \big(1 - \sqrt{\rho}\big)r.$$

Thus we have

$$M^{(t)} \leq \sqrt{\rho} \cdot M^{(t-1)} + \zeta \leq \sqrt{\rho} \cdot r + r(1 - \sqrt{\rho}) \leq r.$$

This completes the proof that for all $t \geq 1$, we have $\mathbf{W}^{(t)} \in \mathbb{B}_F(\mathbf{W}^*; r)$, $\mathbf{\Omega}^{(t)} \in \mathbb{B}_F(\mathbf{\Omega}^*; r)$. Furthermore, we have

$$M^{(t)} \leq \sqrt{\rho} \cdot M^{(t-1)} + \zeta \leq \rho \cdot M^{(t-2)} + \sqrt{\rho} \cdot \zeta + \zeta \leq ...$$
$$\leq \rho^{t/2} \cdot M^{(0)} + \rho^{(t-1)/2} \cdot \zeta + ... + \zeta$$
$$\leq \rho^{t/2} \cdot M^{(0)} + \frac{1}{1 - \sqrt{\rho}}\zeta,$$

where the last inequality holds for series summation rule when $t \to \infty$. Since $M^{(0)} = \max\{\|\mathbf{W}^{(0)} - \mathbf{W}^*\|_F, \|\mathbf{\Omega}^{(0)} - \mathbf{\Omega}^*\|_F\} \leq r$, we rewrite the above inequality as

$$\max\{\|\mathbf{W}^{(t)} - \mathbf{W}^*\|_F, \|\mathbf{\Omega}^{(t)} - \mathbf{\Omega}^*\|_F\} \leq \rho^{t/2} \cdot r + \frac{1}{1-\sqrt{\rho}}\zeta, \text{ for all } t \in [T].$$

This completes the proof. □

## 6 Experiments

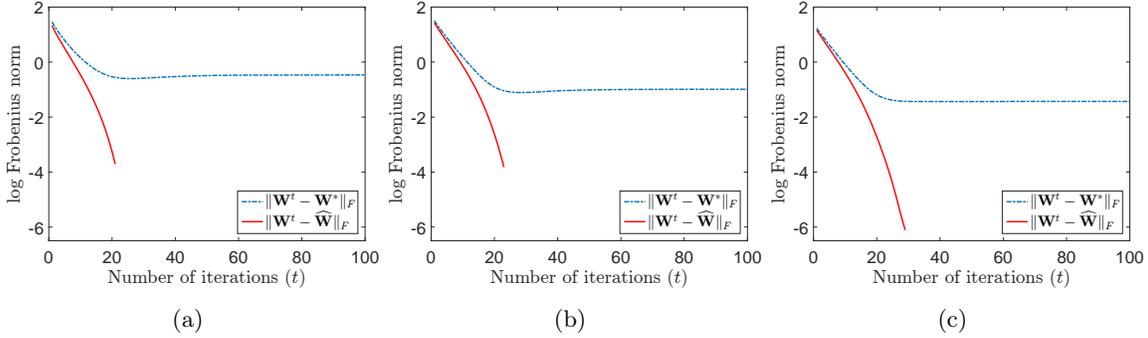

**Figure 1:** Estimation error carves for $\mathbf{W}^*$ in three different scale settings: (a) $n = 1000, m = 10, d = 2000$; (b) $n = 2500, m = 20, d = 5000$ and (c) $n = 5000, m = 30, d = 10000$. $\mathbf{\Omega}^*$ is generated from Hub graph.

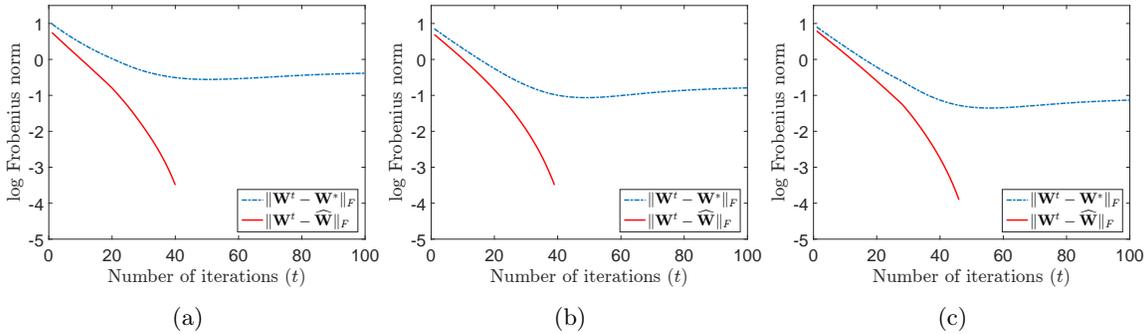

**Figure 2:** Estimation error carves for $\mathbf{W}^*$ in three different scale settings: (a) $n = 1000, m = 10, d = 2000$; (b) $n = 2500, m = 20, d = 5000$ and (c) $n = 5000, m = 30, d = 10000$. $\mathbf{\Omega}^*$ is generated from Band graph.

In this section, we will present numerical results on both synthetic and real datasets to verify the performance of the proposed algorithm in Algorithm 1. We compare our algorithm with several state-of-the-art baseline algorithms: (1) multivariate Lasso (denote as mLasso), (2) multivariate regression with covariance estimation (MRCE) by Rothman et al. (2010); Lee and Liu (2012), and (3) sparse conditional Gaussian graphical models (SCGGM) by Sohn and Kim (2012). Note that MRCE is an alternating minimization based method, which performs much better than the two-step approaches (Lee and Liu, 2012).

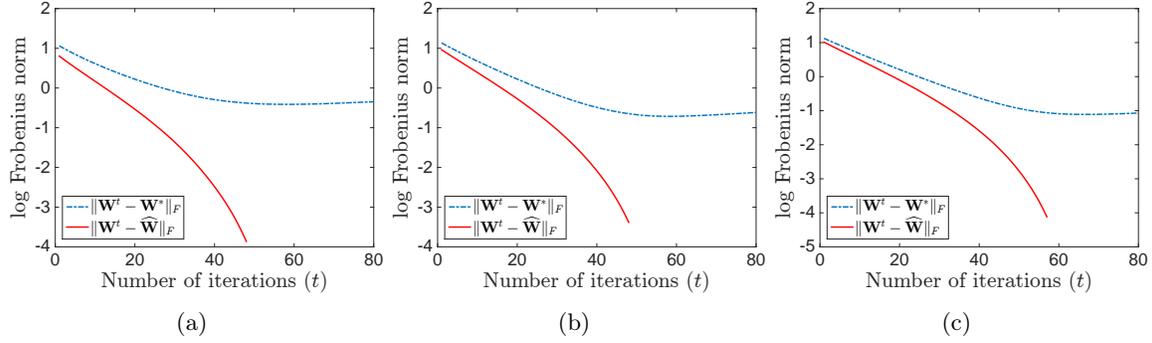

**Figure 3:** Estimation error carves for $\mathbf{W}^*$ in three different scale settings: (a) $n = 1000, m = 10, d = 2000$; (b) $n = 2500, m = 20, d = 5000$ and (c) $n = 5000, m = 30, d = 10000$. $\mathbf{\Omega}^*$ is generated from Scale-free graph.

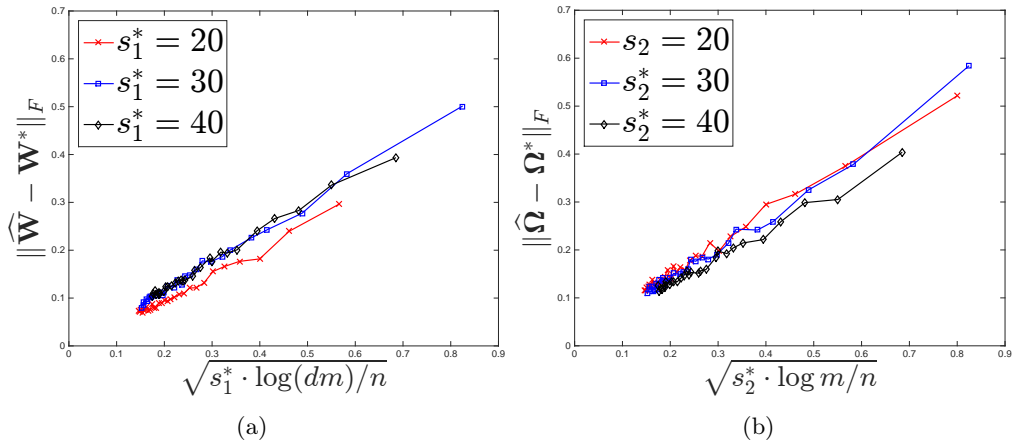

**Figure 4:** (a): Scaled error plot for $\|\widehat{\mathbf{W}} - \mathbf{W}^*\|_F$ under different sparsity settings. $\mathbf{\Omega}^*$ is generated from Band graph. (b): Scaled error plot for $\|\widehat{\mathbf{\Omega}} - \mathbf{\Omega}^*\|_F$ under different sparsity settings. $\mathbf{\Omega}^*$ is generated from Band graph.

### 6.1 Synthetic Data

In each replication of each model, we generate an $n \times d$ predictor matrix $\mathbf{X}$ with rows drawn independently from a multivariate normal distribution $N(\mathbf{0}, \mathbf{\Sigma}_X^*)$, where each element of $\mathbf{\Sigma}_X^*$ is chosen as $[\mathbf{\Sigma}_X^*]_{ij} = 0.6^{|i-j|}$. Similar model has been used by Yuan et al. (2007); Peng et al. (2010); Rothman et al. (2010). On the other hand, each row of the error matrix $\mathbf{E}$ is generated independently from another multivariate normal distribution $N(\mathbf{0}, \mathbf{\Sigma}^*)$ where the precision matrix $\mathbf{\Omega}^* = \mathbf{\Sigma}^{*-1}$ is generated using the following sparse models:

**Band graph**: $\mathbf{\Omega}^* = (\Omega_{ij})$ where $\Omega_{ii} = 1$, $\Omega_{i,i+1} = \Omega_{i+1,i} = 0.4$ and $\Omega_{ij} = 0$ for $|i-j| > 1$.

**Hub graph**: The row/columns are evenly partitioned into $g$ disjoint groups. Each group is associated with a "center" row $i$ in that group. Each pair of off-diagonal elements are set to be related for $i \neq j$ if $j$ also belongs to the same group as $i$ and 0 otherwise. It results in $d - g$ edges in the graph.

**Scale-free graph**: The graph is generated using Barabási Albert algorithm. It results in $d$

edges in the graph $\mathbf{\Omega}'$. To obtain a positive definite precision matrix, the smallest eigenvalue of the $\mathbf{\Omega}'$ (denoted by $e$) is computed. The precision matrix is set to be $\mathbf{\Omega}^* = \mathbf{\Omega}' + e\mathbf{I}$.

We compare the performance on synthetic data with three different settings: (1) $n = 1000, m = 10, d = 2000$ (2) $n = 2500, m = 20, d = 5000$ (3) $n = 5000, m = 30, d = 10000$. Each synthetic dataset is randomly split into the training (50%) and testing set (50%). Each setting is repeated for 10 times. The averaged estimation error, prediction error and running time are reported in Table 1. We can observe that even with growing $m$ and $d$, our algorithm still achieves a decrease in the estimation error of $\mathbf{W}^*$ due to the increase in $n$, while other baseline algorithms cannot. More importantly, our algorithm is much faster than MRCE and SCGGM, especially when the problem size is large (i.e., big $n$ and $d$). The estimation error carves for $\mathbf{W}^*$ under different scale settings and sparse models are demonstrated in Figures 1, 2, 3. As we can see, the optimization error (red solid curve) of our algorithm decreases to zero at a linear rate, while the overall estimator error (blue dashed line) converges to certain level, which is exactly the level of statistical error. Figure 4 illustrates the scaling of the estimation error for $\mathbf{W}^*$ and $\mathbf{\Omega}^*$ respectively. The $x$ axis of these graphs is the rescaled sample size. This result backed up our conclusion that our estimator by Algorithm 1 achieves $O_P\big(\sqrt{s_1^* \log(dm)/n}\big)$ statistical estimation error for $\mathbf{W}^*$, and $O_P\big(\sqrt{s_2^* \log m/n}\big)$ statistical estimation error for $\mathbf{\Omega}^*$.

## 6.2 S&P 500 Stock Price Data

In this experiment, we applied our proposed method to daily closing prices of 452 stocks in the S&P 500 index between January 1, 2003 through January 1, 2008 from Yahoo! Finance (finance.yahoo.com). In order to better characterize the relationship inside the stock data, we use the transformed data, where we calculate the log-ratio of the price at day $t$ to the price at day $t-1$:

$$Y_{t,j} = \log(P_{t+1,j}/P_{t,j}), t = 1, \ldots, T - 1,$$

where $P_{t,j}$ stands for the original price for stock $j$ at day $t$. Here the model we use is a first-order auto-regressive model: $\mathbf{y}_t = \mathbf{W}^{*\top}\mathbf{y}_{t-1} + \boldsymbol{\epsilon}_t$, where $\mathbf{y}_t$ represents the transformed stock prices at time $t$. We randomly generate 10 datasets from the stock data, each of which contains a training set and a testing set. We make sure the time of the training data is ahead of time of the testing data in each dataset.

**Table 1:** Comparison of estimation error (in terms of $\|\widehat{\mathbf{W}} - \mathbf{W}^*\|_F$), prediction error (mean square error) and training time (seconds) on synthetic dataset over 10 replications. N/A means the algorithm cannot output the solution in an hour. $\mathbf{\Omega}^*$ is generated from Hub graph.

|          | n = 1000, m = 10, d = 2000 | | | n = 2500, m = 20, d = 5000 | | | n = 5000, m = 30, d = 10000 | | |
|----------|---------|---------|------|---------|----------|-------|----------|----------|-------|
| Methods  | Est. Err | Pred. Err | Time | Est. Err | Pred. Err | Time | Est. Err | Pred. Err | Time |
| mLasso   | 1.21±0.03 | 6.44±0.14 | 1.9 | 1.46±0.02 | 12.05±0.12 | 10.8 | 1.89±0.01 | 18.60±0.10 | 77.2 |
| MRCE     | 1.14±0.05 | 6.28±0.13 | 3.5 | 1.41±0.03 | 11.87±0.11 | 138.1 | N/A | N/A | N/A |
| SCGGM    | 0.79±0.02 | 5.60±0.12 | 4.6 | 1.35±0.03 | 11.83±0.14 | 20.2 | 1.77±0.01 | 18.16±0.06 | 223.2 |
| Ours     | 0.38±0.02 | 5.10±0.10 | 2.2 | 0.24±0.01 | 10.20±0.09 | 14.1 | 0.21±0.01 | 14.96±0.06 | 92.4 |

| Methods | Prediction Error | Time |
|---------|------------------|------|
| mLasso  | $0.3553 \pm 0.0090$ | 2.5  |
| MRCE    | $0.3400 \pm 0.0079$ | 35.7 |
| SCGGM   | $0.3157 \pm 0.0070$ | 15.0 |
| Ours    | $0.2424 \pm 0.0060$ | 7.3  |

Table 2: Comparison of prediction error (mean square error), and training time (in second) on S&P 500 dataset

| Methods | Prediction Error | Number of Genes | Time |
|---------|------------------|-----------------|------|
| mLasso  | $1.21 \pm 0.10$  | $180.8 \pm 8.4$  | 0.16 |
| MRCE    | $1.20 \pm 0.07$  | $162.6 \pm 14.2$ | 1.15 |
| SCGGM   | $1.32 \pm 0.07$  | $394.2 \pm 7.8$  | 0.23 |
| aMCR    | $1.19 \pm 0.01$  | $65.2 \pm 1.7$   | -    |
| Ours    | $1.09 \pm 0.10$  | $234.0 \pm 8.6$  | 0.18 |

Table 3: Comparison of prediction error (mean square error), and training time (in second) on GBM cancer dataset

As can be seen in Table 2, our proposed algorithm outperforms all the state-of-the-art baselines overwhelmingly in terms of prediction error. It achieves major improvement in prediction error (around 30% performance improvement over mLasso) without paying too much computational cost. In contrast, MRCE and SCGGM only achieve a little improvement over mLasso in prediction accuracy while costing a lot more computational time.

### 6.3 Glioblastoma Multiforme (GBM) Cancer Data

In this experiment, we applied the proposed method to a Glioblastoma multiforme (GBM) cancer dataset studied by the Cancer Genome Atlas (TCGA) Research Network (McLendon et al., 2008; Verhaak et al., 2010; Wang, 2013). One of our primary goal is to study the relationship between the microRNA expressions and gene expressions, revealed by the sparse coefficient matrix. It is also of interest to construct the underlying network among the microRNAs, which is interpreted by the structure of the sparse precision matrix.

For illustration, some preliminary data cleaning is conducted as in McLendon et al. (2008) and Lee and Liu (2012). Furthermore, the top 500 genes and top 20 microRNAs with large MADs (median absolute deviation) are selected. The dataset is randomly split into a training set with 120 samples and a test set with 76 samples. The estimation performance is measured by the mean square error. Besides comparing with the three baseline algorithms, we also compare with adapted multivariate conditional regression algorithm (aMCR) proposed by Wang (2013), since we use the same dataset.

Table 3 demonstrates the results for five different methods including ours. Our proposed algorithm achieves the best prediction accuracy among the state-of-the-art algorithms. The experimental result for aMCR is gathered directly from Wang (2013) (They do not report the training time). We can see that our algorithm outperforms all the baseline algorithms including aMCR by a large margin. Since the size of this dataset is moderate, the training time of MRCE, SCGGM and our algorithm are comparable. We also note that our algorithm selects more genes than MRCE and aMCR, but fewer genes than SCGGM.

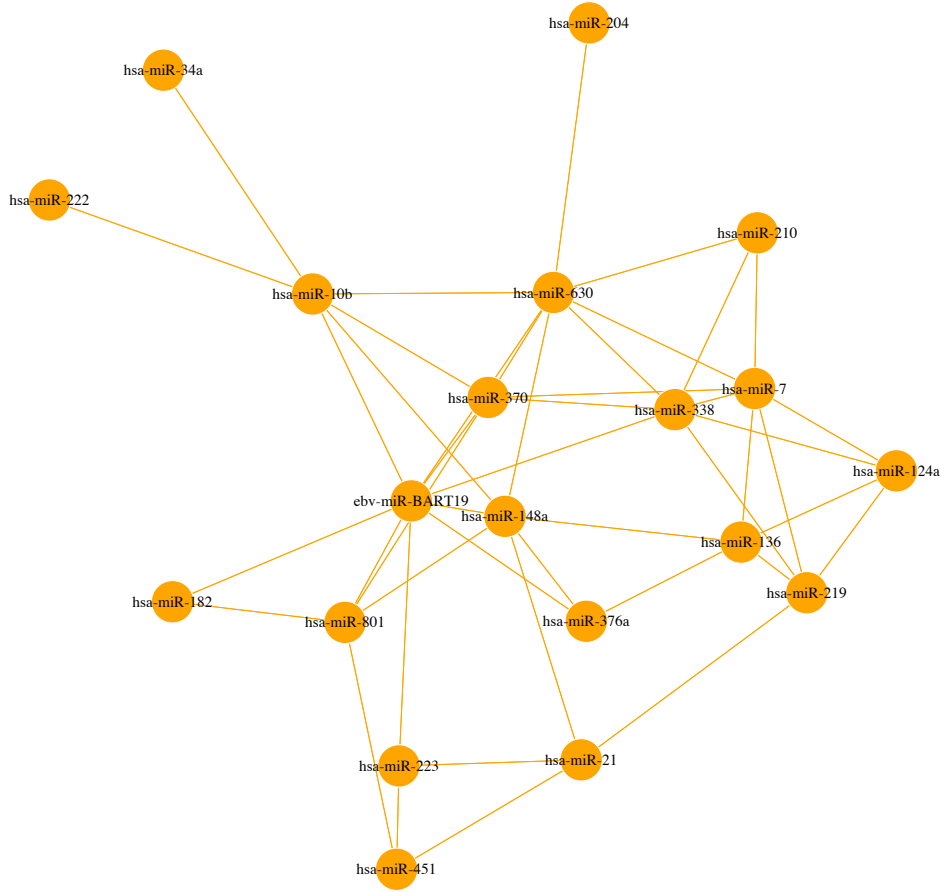

**Figure 5:** Graphical dependency network of selected microRNAs based on estimated sparse precision matrix.

Figure 5 illustrates the dependency network of the 20 microRNAs corresponding to the estimated sparse precision matrix. We can clearly observe that the microRNA expression levels for ebv-miR-BART19, hsa-miR-370, hsa-miR-148a, hsa-miR-338 are highly dependent on the existence of other microRNAs, while for microRNAs on the outer sphere of Figure 5, such as hsa-miR-204, hsa-miR-34a, hsa-miR-222, their expression levels are relatively independent from other microRNAs'. Our observation is consistent with the results reported in Lee and Liu (2012); Wang (2013).

# 7 Conclusions

In this paper, we presented a gradient descent algorithm with hard thresholding for joint multivariate regression and precision matrix estimation in the high dimensional regime, under cardinality constraints. It attains a linear convergence to the true regression coefficients and precision matrix simultaneously, up to a near optimal statistical error. Thorough experiments on both synthetic and real datasets backup our theory.

# A  Proof of Technical Lemmas in Section 5

In the following, we will give detailed proof of the technical lemmas used in Section 5.

## A.1  Proof of Lemma 5.1

*Proof.* Recall that

$$f_n(\mathbf{W}, \mathbf{\Omega}) = -\log|\mathbf{\Omega}| + \frac{1}{n}\sum_{i=1}^{n}\left((\mathbf{y}_i - \mathbf{W}^\top \mathbf{x}_i)^\top \mathbf{\Omega}(\mathbf{y}_i - \mathbf{W}^\top \mathbf{x}_i)\right)$$

$$= -\log|\mathbf{\Omega}| + \frac{1}{n}\sum_{i=1}^{n}\left((\mathbf{W}^{*\top}\mathbf{x}_i - \mathbf{W}^\top \mathbf{x}_i + \boldsymbol{\epsilon}_i)^\top \mathbf{\Omega}(\mathbf{W}^{*\top}\mathbf{x}_i - \mathbf{W}^\top \mathbf{x}_i + \boldsymbol{\epsilon}_i)\right). \quad (\text{A.1})$$

Based on the above equality we compute the population version of $f$ function:

$$f(\mathbf{W}, \mathbf{\Omega}) = \mathbb{E}\big[f_n(\mathbf{W}, \mathbf{\Omega})\big]$$

$$= \mathbb{E}\left[-\log|\mathbf{\Omega}| + \frac{1}{n}\sum_{i=1}^{n}\left((\mathbf{W}^{*\top}\mathbf{x}_i - \mathbf{W}^\top \mathbf{x}_i + \boldsymbol{\epsilon}_i)^\top \mathbf{\Omega}(\mathbf{W}^{*\top}\mathbf{x}_i - \mathbf{W}^\top \mathbf{x}_i + \boldsymbol{\epsilon}_i)\right)\right]$$

$$= -\log|\mathbf{\Omega}| + \frac{1}{n}\sum_{i=1}^{n}\mathbf{x}_i^\top(\mathbf{W}^* - \mathbf{W})\mathbf{\Omega}(\mathbf{W}^* - \mathbf{W})^\top \mathbf{x}_i + \text{tr}(\mathbf{\Omega}\mathbf{\Omega}^{*-1}). \quad (\text{A.2})$$

Thus, we get

$$\nabla_1 f(\mathbf{W}, \mathbf{\Omega}) = -\frac{2}{n}\sum_{i=1}^{n}\mathbf{x}_i\mathbf{x}_i^\top(\mathbf{W}^* - \mathbf{W})\mathbf{\Omega}. \quad (\text{A.3})$$

Do vectorization and use the property of Kronecker product that $\text{vec}(\mathbf{ABC}) = (\mathbf{C}^\top \otimes \mathbf{A})\text{vec}(\mathbf{B})$, we obtain

$$\nabla^2_{\text{vec}(\mathbf{W})} f(\mathbf{W}, \mathbf{\Omega}^*) = \mathbf{\Omega}^* \otimes \left(\frac{2}{n}\mathbf{X}^\top \mathbf{X}\right) = 2\mathbf{\Omega}^* \otimes \widehat{\mathbf{\Sigma}}_X.$$

For function $f(\cdot, \mathbf{\Omega}^*)$, according to Taylor expansion, we have

$$f(\mathbf{W}', \mathbf{\Omega}^*) = f(\mathbf{W}, \mathbf{\Omega}^*) + \langle \nabla_1 f(\mathbf{W}, \mathbf{\Omega}^*), \mathbf{W}' - \mathbf{W}\rangle$$

$$+ \frac{1}{2}\langle \text{vec}(\mathbf{W}') - \text{vec}(\mathbf{W}), \nabla^2_{\text{vec}(\mathbf{W})} f(\mathbf{W}, \mathbf{\Omega}^*)(\text{vec}(\mathbf{W}') - \text{vec}(\mathbf{W}))\rangle,$$

which further implies

$$f(\mathbf{W}', \mathbf{\Omega}^*) - f(\mathbf{W}, \mathbf{\Omega}^*) - \langle \nabla_1 f(\mathbf{W}, \mathbf{\Omega}^*), \mathbf{W}' - \mathbf{W}\rangle \le \frac{1}{2}\lambda_{\max}(\nabla^2_{\text{vec}(\mathbf{W})} f(\mathbf{W}, \mathbf{\Omega}^*))\|\mathbf{W}' - \mathbf{W}\|_F^2,$$

$$f(\mathbf{W}', \mathbf{\Omega}^*) - f(\mathbf{W}, \mathbf{\Omega}^*) - \langle \nabla_1 f(\mathbf{W}, \mathbf{\Omega}^*), \mathbf{W}' - \mathbf{W}\rangle \ge \frac{1}{2}\lambda_{\min}(\nabla^2_{\text{vec}(\mathbf{W})} f(\mathbf{W}, \mathbf{\Omega}^*))\|\mathbf{W}' - \mathbf{W}\|_F^2.$$

Assumption 4.2 immediately implies that $\|\widehat{\boldsymbol{\Sigma}}_X\|_2 \leq 1$. Under Assumptions 4.1 and 4.2, using properties of Kronecker product that $\lambda_{\min}(\mathbf{A} \otimes \mathbf{B}) = \lambda_{\min}(\mathbf{A}) \times \lambda_{\min}(\mathbf{B})$ and $\lambda_{\max}(\mathbf{A} \otimes \mathbf{B}) = \lambda_{\max}(\mathbf{A}) \times \lambda_{\max}(\mathbf{B})$, we have

$$\frac{1}{\nu\tau} \leq \lambda_{\min}(\nabla^2_{\text{vec}(\mathbf{W})} f(\mathbf{W}, \boldsymbol{\Omega}^*)) \leq \lambda_{\max}(\nabla^2_{\text{vec}(\mathbf{W})} f(\mathbf{W}, \boldsymbol{\Omega}^*)) \leq 2\nu.$$

Therefore, function $f(\cdot, \boldsymbol{\Omega}^*)$ is $(1/(\nu\tau))$-strongly convex and $(2\nu)$-smooth function:

$$\frac{1}{2\nu\tau}\|\mathbf{W}' - \mathbf{W}\|_F^2 \leq f(\mathbf{W}', \boldsymbol{\Omega}^*) - f(\mathbf{W}, \boldsymbol{\Omega}^*) - \nabla_1 f(\mathbf{W}, \boldsymbol{\Omega}^*)^\top(\mathbf{W}' - \mathbf{W}) \leq \nu \cdot \|\mathbf{W}' - \mathbf{W}\|_F^2.$$

□

## A.2 Proof of Lemma 5.2

*Proof.* From (A.2), we have

$$\nabla^2_{\text{vec}(\boldsymbol{\Omega})} f(\mathbf{W}^*, \boldsymbol{\Omega}) = \boldsymbol{\Omega}^{-1} \otimes \boldsymbol{\Omega}^{-1}.$$

According to Mean Value Theorem, we have

$$f(\mathbf{W}^*, \boldsymbol{\Omega}') = f(\mathbf{W}^*, \boldsymbol{\Omega}) + \langle \nabla_1 f(\mathbf{W}^*, \boldsymbol{\Omega}), \boldsymbol{\Omega}' - \boldsymbol{\Omega} \rangle$$
$$+ \frac{1}{2}\langle \text{vec}(\boldsymbol{\Omega}') - \text{vec}(\boldsymbol{\Omega}), \nabla^2_{\text{vec}(\boldsymbol{\Omega})} f(\mathbf{W}^*, \mathbf{Z})(\text{vec}(\boldsymbol{\Omega}') - \text{vec}(\boldsymbol{\Omega}))\rangle,$$

where $\mathbf{Z} = t\boldsymbol{\Omega}' + (1-t)\boldsymbol{\Omega}$ with $t \in [0, 1]$. Define $\boldsymbol{\Delta} = \boldsymbol{\Omega}' - \boldsymbol{\Omega}$, we have

$$\lambda_{\min}(\nabla^2_{\text{vec}(\boldsymbol{\Omega})} f(\mathbf{W}^*, \mathbf{Z})) = \lambda_{\min}(\mathbf{Z}^{-1} \otimes \mathbf{Z}^{-1}) = \lambda_{\min}(\mathbf{Z}^{-1})^2 = \|\boldsymbol{\Omega} + t\boldsymbol{\Delta}\|_2^{-2}$$
$$\geq [\|\boldsymbol{\Omega}^*\|_2 + \|\boldsymbol{\Omega} - \boldsymbol{\Omega}^*\|_2 + t\|\boldsymbol{\Delta}\|_2]^{-2}$$
$$\geq [\|\boldsymbol{\Omega}^*\|_F + 3r]^{-2},$$

where the first inequality holds due to triangle inequality, and the second inequality holds for $\|\boldsymbol{\Omega} - \boldsymbol{\Omega}^*\|_2 \leq \|\boldsymbol{\Omega} - \boldsymbol{\Omega}^*\|_F \leq r$ and $t\|\boldsymbol{\Delta}\|_2 \leq \|\boldsymbol{\Delta}\|_2 \leq \|\boldsymbol{\Delta}\|_F \leq \|\boldsymbol{\Omega} - \boldsymbol{\Omega}^*\|_F + \|\boldsymbol{\Omega}' - \boldsymbol{\Omega}^*\|_F \leq 2r$. For the same reason, we have

$$\lambda_{\max}(\nabla^2_{\text{vec}(\boldsymbol{\Omega})} f(\mathbf{W}^*, \mathbf{Z})) \leq [\|\boldsymbol{\Omega}^*\|_F + 3r]^2,$$

Therefore, we have

$$\frac{1}{2[\|\boldsymbol{\Omega}^*\|_F + 3r]^2}\|\boldsymbol{\Omega}' - \boldsymbol{\Omega}\|_F^2 \leq f(\mathbf{W}^*, \boldsymbol{\Omega}') - f(\mathbf{W}^*, \boldsymbol{\Omega}) - \nabla_2 f(\mathbf{W}^*, \boldsymbol{\Omega})^\top(\boldsymbol{\Omega}' - \boldsymbol{\Omega})$$
$$\leq \frac{[\|\boldsymbol{\Omega}^*\|_F + 3r]^2}{2}\|\boldsymbol{\Omega}' - \boldsymbol{\Omega}\|_F^2.$$

This completes the proof. □

## A.3 Proof of Lemma 5.3

*Proof.* First, we bound $\|\nabla_1 f(\mathbf{W}, \mathbf{\Omega}^*) - \nabla_1 f(\mathbf{W}, \mathbf{\Omega})\|_F$ in (5.1). From (A.3) we have

$$\nabla_1 f(\mathbf{W}, \mathbf{\Omega}) = -\frac{2}{n}\sum_{i=1}^n \mathbf{x}_i \mathbf{x}_i^\top (\mathbf{W}^* - \mathbf{W})\mathbf{\Omega},$$

$$\nabla_1 f(\mathbf{W}, \mathbf{\Omega}^*) = -\frac{2}{n}\sum_{i=1}^n \mathbf{x}_i \mathbf{x}_i^\top (\mathbf{W}^* - \mathbf{W})\mathbf{\Omega}^*.$$

Thus we get

$$\begin{aligned}\|\nabla_1 f(\mathbf{W}, \mathbf{\Omega}^*) - \nabla_1 f(\mathbf{W}, \mathbf{\Omega})\|_F &= \left\|\frac{2}{n}\sum_{i=1}^n \mathbf{x}_i \mathbf{x}_i^\top (\mathbf{W}^* - \mathbf{W})(\mathbf{\Omega}^* - \mathbf{\Omega})\right\|_F \\ &\le 2\left\|\frac{1}{n}\sum_{i=1}^n \mathbf{x}_i \mathbf{x}_i^\top\right\|_F \cdot \|\mathbf{W}^* - \mathbf{W}\|_F \cdot \|\mathbf{\Omega}^* - \mathbf{\Omega}\|_F \\ &\le \frac{2}{n}\sum_{i=1}^n \|\mathbf{x}_i \mathbf{x}_i^\top\|_F \cdot \|\mathbf{W}^* - \mathbf{W}\|_F \cdot \|\mathbf{\Omega}^* - \mathbf{\Omega}\|_F,\end{aligned} \quad (A.4)$$

where the first inequality holds due to Cauchy-Schwarz inequality, and the second inequality follows from triangle inequality.

Since we know that $\|\mathbf{W} - \mathbf{W}^*\|_F \le r$, and

$$\|\mathbf{x}_i \mathbf{x}_i^\top\|_F = \sqrt{\mathrm{tr}\left(\mathbf{x}_i \mathbf{x}_i^\top \mathbf{x}_i \mathbf{x}_i^\top\right)} \le \sqrt{\mathrm{tr}\left(\mathbf{x}_i \mathbf{x}_i^\top\right)} = \sqrt{\mathbf{x}_i^\top \mathbf{x}_i} \le 1,$$

thus (A.4) can be further bounded as

$$\|\nabla_1 f(\mathbf{W}, \mathbf{\Omega}^*) - \nabla_1 f(\mathbf{W}, \mathbf{\Omega})\|_F \le 2r \cdot \|\mathbf{\Omega}^* - \mathbf{\Omega}\|_F.$$

Now consider $\|\nabla_2 f(\mathbf{W}^*, \mathbf{\Omega}) - \nabla_2 f(\mathbf{W}, \mathbf{\Omega})\|_F$ in (5.2). From (A.2), we have

$$\nabla_2 f(\mathbf{W}, \mathbf{\Omega}) = -\mathbf{\Omega}^{-1} + \mathbf{\Omega}^{*-1} + \frac{1}{n}\sum_{i=1}^n (\mathbf{W}^* - \mathbf{W})^\top \mathbf{x}_i \mathbf{x}_i^\top (\mathbf{W}^* - \mathbf{W})$$

$$\nabla_2 f(\mathbf{W}^*, \mathbf{\Omega}) = -\mathbf{\Omega}^{-1} + \mathbf{\Omega}^{*-1}.$$

Thus, we obtain

$$\begin{aligned}\|\nabla_2 f(\mathbf{W}^*, \mathbf{\Omega}) - \nabla_2 f(\mathbf{W}, \mathbf{\Omega})\|_F &= \left\|\frac{1}{n}\sum_{i=1}^n (\mathbf{W}^* - \mathbf{W})^\top \mathbf{x}_i \mathbf{x}_i^\top (\mathbf{W}^* - \mathbf{W})\right\|_F \\ &\le \left\|\frac{1}{n}\sum_{i=1}^n \mathbf{x}_i \mathbf{x}_i^\top\right\|_F \cdot \|\mathbf{W}^* - \mathbf{W}\|_F^2,\end{aligned}$$

where the inequality follows from Cauchy-Schwarz inequality. Following similar proof procedure in (A.4), we can further bound the above inequality as

$$\|\nabla_2 f(\mathbf{W}^*, \mathbf{\Omega}) - \nabla_2 f(\mathbf{W}, \mathbf{\Omega})\|_F \le \frac{1}{n}\sum_{i=1}^n \|\mathbf{x}_i \mathbf{x}_i^\top\|_F \cdot \|\mathbf{W}^* - \mathbf{W}\|_F^2 \le r \cdot \|\mathbf{W}^* - \mathbf{W}\|_F.$$

□

## A.4 Proof of Lemma 5.4

*Proof.* To simplify the proof, here we prove a more general case by substituting $n/T$ with $n$ and substituting $\delta/T$ with $\delta$. Since $\boldsymbol{\epsilon}_i \sim N(\mathbf{0}, \boldsymbol{\Sigma}^*)$, we have $\max_{ij}\|\epsilon_{ij}\|_{\psi_2} \leq C_1 \lambda_{\max}(\boldsymbol{\Sigma}^*) \leq C_1 \nu$. Recall that

$$f_n(\mathbf{W}, \boldsymbol{\Omega}) = -\log|\boldsymbol{\Omega}| + \frac{1}{n}\sum_{i=1}^{n}\left((\mathbf{W}^{*\top}\mathbf{x}_i - \mathbf{W}^{\top}\mathbf{x}_i + \boldsymbol{\epsilon}_i)^{\top}\boldsymbol{\Omega}(\mathbf{W}^{*\top}\mathbf{x}_i - \mathbf{W}^{\top}\mathbf{x}_i + \boldsymbol{\epsilon}_i)\right). \quad (A.5)$$

First we prove the bound in (5.3). From (A.1), we get

$$\nabla_1 f_n(\mathbf{W}, \boldsymbol{\Omega}) = -\frac{2}{n}\sum_{i=1}^{n}\mathbf{x}_i\mathbf{x}_i^{\top}(\mathbf{W}^* - \mathbf{W})\boldsymbol{\Omega} - \frac{2}{n}\sum_{i=1}^{n}\mathbf{x}_i\boldsymbol{\epsilon}_i^{\top}\boldsymbol{\Omega}. \quad (A.6)$$

From (A.2) we have

$$\nabla_1 f(\mathbf{W}, \boldsymbol{\Omega}) = -\frac{2}{n}\sum_{i=1}^{n}\mathbf{x}_i\mathbf{x}_i^{\top}(\mathbf{W}^* - \mathbf{W})\boldsymbol{\Omega}. \quad (A.7)$$

Combining (A.6) and (A.7) we obtain

$$\nabla_1 f(\mathbf{W}, \boldsymbol{\Omega}) - \nabla_1 f_n(\mathbf{W}, \boldsymbol{\Omega}) = \frac{2}{n}\sum_{i=1}^{n}\mathbf{x}_i\boldsymbol{\epsilon}_i^{\top}\boldsymbol{\Omega}. \quad (A.8)$$

In the following, let $\mathbf{A}^{(i)} = \mathbf{x}_i\boldsymbol{\epsilon}_i^{\top}\boldsymbol{\Omega}$. Consider the $\psi_2$ norm of each element in $\mathbf{A}^{(i)}$,

$$\|A_{jk}^{(i)}\|_{\psi_2}^2 = \left\|x_{ij}\sum_{\ell=1}^{m}\epsilon_{i\ell}\Omega_{\ell k}\right\|_{\psi_2}^2 \leq \left\|\sum_{\ell=1}^{m}\epsilon_{i\ell}\Omega_{\ell k}\right\|_{\psi_2}^2 \leq C_2\sum_{\ell=1}^{m}\Omega_{\ell k}^2\|\epsilon_{i\ell}\|_{\psi_2}^2, \quad (A.9)$$

where the first inequality holds for the fact that $\|\mathbf{x}_i\|_2 \leq 1$, and the second inequality holds for the rotation invariance property of $\psi_2$ norm in Lemma C.1. Since $\max_{ij}\|\epsilon_{ij}\|_{\psi_2} \leq C_1 \nu$, (A.9) can be further bounded by

$$\|A_{jk}^{(i)}\|_{\psi_2}^2 \leq C_2\sum_{\ell=1}^{m}\Omega_{\ell k}^2 C_1^2 \nu^2 \leq C_3\|\boldsymbol{\Omega}\|_F^2 \nu^2 \leq C_3(r + \|\boldsymbol{\Omega}^*\|_F)^2 \nu^2.$$

By Hoeffding-type inequality in Theorem C.4, we have

$$\mathbb{P}\left\{\left|\frac{2}{n}\sum_{i=1}^{n}A_{jk}^{(i)}\right| \geq t\right\} \leq \exp\left(-\frac{C_4 t^2 n}{4C_3(r + \|\boldsymbol{\Omega}^*\|_F)^2 \nu^2}\right).$$

Applying union bound to all possible pairs of $j \in [d]$, $k \in [m]$, we get

$$\mathbb{P}\left\{\|\nabla_1 f(\mathbf{W}, \boldsymbol{\Omega}) - \nabla_1 f_n(\mathbf{W}, \boldsymbol{\Omega})\|_{\infty,\infty} \geq t\right\} \leq \sum_{j,k}\mathbb{P}\left\{\left|\frac{2}{n}\sum_{i=1}^{n}A_{jk}^{(i)}\right| \geq t\right\}$$

$$\leq d \cdot m \cdot \exp\left(-\frac{C_4 t^2 n}{4C_3(r + \|\boldsymbol{\Omega}^*\|_F)^2 \nu^2}\right).$$

Choose $t = 2\nu(r + \|\mathbf{\Omega}^*\|_F)\sqrt{C_3/C_4}\sqrt{(2\log d + \log m)/n}$ and $C = 2\sqrt{C_3/C_4}$, with probability at least $1 - 1/d$ we have

$$\|\nabla_1 f(\mathbf{W}, \mathbf{\Omega}) - \nabla_1 f_n(\mathbf{W}, \mathbf{\Omega})\|_{\infty,\infty} \leq C(r + \|\mathbf{\Omega}^*\|_F)\nu\sqrt{\frac{2\log d + \log m}{n}},$$

which immediately implies the conclusion in (5.3). Now we give the bound for (5.4). From (A.1), we have

$$\nabla_2 f_n(\mathbf{W}, \mathbf{\Omega}) = -\mathbf{\Omega}^{-1} + \frac{1}{n}\sum_{i=1}^{n}(\mathbf{W}^{*\top}\mathbf{x}_i - \mathbf{W}^{\top}\mathbf{x}_i + \boldsymbol{\epsilon}_i)(\mathbf{W}^{*\top}\mathbf{x}_i - \mathbf{W}^{\top}\mathbf{x}_i + \boldsymbol{\epsilon}_i)^{\top}. \tag{A.10}$$

From (A.2) we obtain

$$\nabla_2 f(\mathbf{W}, \mathbf{\Omega}) = -\mathbf{\Omega}^{-1} + \frac{1}{n}\sum_{i=1}^{n}(\mathbf{W}^* - \mathbf{W})^{\top}\mathbf{x}_i\mathbf{x}_i^{\top}(\mathbf{W}^* - \mathbf{W}) + \mathbf{\Omega}^{*-1}. \tag{A.11}$$

Thus we get

$$\nabla_2 f(\mathbf{W}, \mathbf{\Omega}) - \nabla_2 f_n(\mathbf{W}, \mathbf{\Omega}) = -\frac{1}{n}\sum_{i=1}^{n}\boldsymbol{\epsilon}_i\boldsymbol{\epsilon}_i^{\top} + \mathbf{\Omega}^{*-1} - \frac{1}{n}\sum_{i=1}^{n}(\mathbf{W}^* - \mathbf{W})^{\top}\mathbf{x}_i\boldsymbol{\epsilon}_i^{\top}$$
$$- \frac{1}{n}\sum_{i=1}^{n}\boldsymbol{\epsilon}_i\mathbf{x}_i^{\top}(\mathbf{W}^* - \mathbf{W}).$$

Then we have

$$\|\nabla_2 f(\mathbf{W}, \mathbf{\Omega}) - \nabla_2 f_n(\mathbf{W}, \mathbf{\Omega})\|_{\infty,\infty} \leq \underbrace{\left\|\frac{1}{n}\sum_{i=1}^{n}\boldsymbol{\epsilon}_i\boldsymbol{\epsilon}_i^{\top} - \boldsymbol{\Sigma}^*\right\|_{\infty,\infty}}_{I_1} + \underbrace{\left\|\frac{1}{n}\sum_{i=1}^{n}(\mathbf{W}^* - \mathbf{W})^{\top}\mathbf{x}_i\boldsymbol{\epsilon}_i^{\top}\right\|_{\infty,\infty}}_{I_2}$$
$$+ \underbrace{\left\|\frac{1}{n}\sum_{i=1}^{n}\boldsymbol{\epsilon}_i\mathbf{x}_i^{\top}(\mathbf{W}^* - \mathbf{W})\right\|_{\infty,\infty}}_{I_3}, \tag{A.12}$$

In the following proof, let $\mathbf{C}^{(i)} = (\mathbf{W}^* - \mathbf{W})^{\top}\mathbf{x}_i\boldsymbol{\epsilon}_i^{\top}$ and $\mathbf{D}^{(i)} = \boldsymbol{\epsilon}_i\mathbf{x}_i^{\top}(\mathbf{W}^* - \mathbf{W})$. For term $I_1$, by Lemma C.7, we have, with probability at least $1 - 4/n^2$

$$\left\|\frac{1}{n}\sum_{i=1}^{n}\boldsymbol{\epsilon}_i\boldsymbol{\epsilon}_i^{\top} - \boldsymbol{\Sigma}^*\right\|_{\infty,\infty} \leq C_6\nu\sqrt{\frac{\log m}{n}}. \tag{A.13}$$

For term $I_2$, consider each element in the matrix, i.e., $C_{jk}^{(i)}$, we have

$$\|C_{jk}^{(i)}\|_{\psi_2}^2 = \left\|\sum_{\ell=1}^{d}(W_{\ell j}^* - W_{\ell j})x_{i\ell}\epsilon_{ik}\right\|_{\psi_2}^2 \leq C_2\sum_{\ell=1}^{d}(W_{\ell j}^* - W_{\ell j})^2 x_{i\ell}^2\|\epsilon_{i\ell}\|_{\psi_2}^2$$
$$\leq C_2\sum_{\ell=1}^{d}(W_{\ell j}^* - W_{\ell j})^2 \sum_{\ell=1}^{d}x_{i\ell}^2\|\epsilon_{i\ell}\|_{\psi_2}^2 \leq C_2\|\mathbf{W}^* - \mathbf{W}\|_F^2 C_1^2\nu^2 \leq C_2 r^2 C_1^2 \nu^2.$$

By Hoeffding-type inequality in Theorem C.4, we have

$$\mathbb{P}\left\{\left|\frac{1}{n}\sum_{i=1}^{n}C_{jk}^{(i)}\right| \geq t\right\} \leq \exp\left(-\frac{C_4 t^2 n}{C_2 r^2 C_1^2 \nu^2}\right).$$

Apply union bound to all possible pairs of $j \in [m]$, $k \in [m]$, we get

$$\mathbb{P}\left\{\left\|\frac{1}{n}\sum_{i=1}^{n}(\mathbf{W}^* - \mathbf{W})^\top \mathbf{x}_i \boldsymbol{\epsilon}_i^\top\right\|_{\infty,\infty} \geq t\right\} \leq m^2 \exp\left(-\frac{C_4 t^2 n}{C_2 r^2 C_1^2 \nu^2}\right).$$

Choose $t = r\nu C_7 \sqrt{2 \log m / n}$ and $C_7 = \sqrt{C_2/C_4}C_1$ with probability at least $1 - 1/d$ we have that

$$\left\|\frac{1}{n}\sum_{i=1}^{n}(\mathbf{W}^* - \mathbf{W})^\top \mathbf{x}_i \boldsymbol{\epsilon}_i^\top\right\|_{\infty,\infty} \leq r\nu C_7 \sqrt{\frac{\log m}{n}}. \tag{A.14}$$

For term $I_3$, since $\mathbf{D}^{(i)} = \mathbf{C}^{(i)\top}$, it holds the same conclusion for term $I_3$ that, with probability at least $1 - 1/d$ we have

$$\left\|\frac{1}{n}\sum_{i=1}^{n}\boldsymbol{\epsilon}_i \mathbf{x}_i^\top (\mathbf{W}^* - \mathbf{W})\right\|_{\infty,\infty} \leq r\nu C_7 \sqrt{\frac{\log m}{n}}. \tag{A.15}$$

Submit (A.13),(A.14), (A.15) into (A.12) and apply union bound we have, with probability at least $1 - 2/d - 4/n^2$, that

$$\left\|\nabla_2 f(\mathbf{W}, \boldsymbol{\Omega}) - \nabla_2 f_n(\mathbf{W}, \boldsymbol{\Omega})\right\|_{\infty,\infty} \leq C' r\nu \sqrt{\frac{\log m}{n}}.$$

It immediately implies the conclusion in (5.4) and this completes the proof. $\square$

### A.5 Proof of Lemma 5.5

In order to prove Lemma 5.5, we need the following auxiliary lemma.

**Lemma A.1** (Nesterov (2004)). Under Assumptions 4.1 and 4.2, let $\eta = 2/(L_1 + \mu_1)$, suppose $\mathbf{W}^+$ is obtained by the following gradient descent update form

$$\mathbf{W}^+ = \mathbf{W} - \eta \nabla_1 f(\mathbf{W}, \boldsymbol{\Omega}^*).$$

We have

$$\|\mathbf{W}^+ - \mathbf{W}^*\|_F \leq \frac{L_1 - \mu_1}{L_1 + \mu_1}\|\mathbf{W} - \mathbf{W}^*\|_F. \tag{A.16}$$

*Proof of Lemma 5.5.* For notation simplicity, let $\mathbf{W}^+$ stands for $\overline{\mathbf{W}}^{(t+0.5)}$, $\mathbf{W}$ stands for $\mathbf{W}^{(t)}$ and $\boldsymbol{\Omega}$ stands for $\boldsymbol{\Omega}^{(t)}$. Also let $L_1 = 2\nu$, $\mu_1 = 1/\nu\tau$, $\eta_1 = 2\nu\tau/(2\nu^2\tau + 1)$ and $\gamma_1 = 2r$. We have

$$\begin{aligned}\|\mathbf{W}^+ - \mathbf{W}^*\|_F &= \|\mathbf{W} - \eta_1 \nabla_1 f(\mathbf{W}, \boldsymbol{\Omega}) - \mathbf{W}^*\|_F \\ &\leq \|\mathbf{W} - \eta_1 \nabla_1 f(\mathbf{W}, \boldsymbol{\Omega}^*) - \mathbf{W}^*\|_F + \eta_1 \|\nabla_1 f(\mathbf{W}, \boldsymbol{\Omega}) - \nabla_1 f(\mathbf{W}, \boldsymbol{\Omega}^*)\|_F \end{aligned} \tag{A.17}$$

where the inequality holds due to triangle inequality. Submit the conclusion (A.16) in Lemma A.1 into the above equality, we obtain

$$\|\mathbf{W}^+ - \mathbf{W}^*\|_F \leq \frac{L_1 - \mu_1}{L_1 + \mu_1}\|\mathbf{W} - \mathbf{W}^*\|_F + \frac{2\gamma_1}{L_1 + \mu_1} \cdot \|\mathbf{\Omega} - \mathbf{\Omega}^*\|_F, \quad (A.18)$$

where the last term on the right side of the above inequality follows from Lemma 5.3, in which we obtain $\|\nabla_1 f(\mathbf{W}, \mathbf{\Omega}^*) - \nabla_1 f(\mathbf{W}, \mathbf{\Omega})\|_F \leq \gamma_1 \cdot \|\mathbf{\Omega}^* - \mathbf{\Omega}\|_F$. By submitting the definition of $L_1$, $\mu_1$, $\eta_1$ and $\gamma_1$ back into (A.18) we complete the proof. □

## A.6 Proof of Lemma 5.6

We omit the proof since it is similar to the proof of Lemma 5.5.

## A.7 Proof of Lemma 5.8

*Proof.* First we consider the left hand side in (5.9):

$$\begin{aligned}
\left\|\mathrm{HT}(\mathbf{W}^{(t+0.5)}, \widehat{S}_1^{(t+0.5)}) - \mathrm{HT}(\overline{\mathbf{W}}^{(t+0.5)}, \widehat{S}_1^{(t+0.5)})\right\|_F &= \left\|\left[\mathbf{W}^{(t+0.5)} - \overline{\mathbf{W}}^{(t+0.5)}\right]_{\widehat{S}_1^{(t+0.5)}}\right\|_F \\
&\leq \sqrt{s_1} \cdot \left\|\left[\mathbf{W}^{(t+0.5)} - \overline{\mathbf{W}}^{(t+0.5)}\right]_{\widehat{S}_1^{(t+0.5)}}\right\|_{\infty,\infty} \\
&\leq \sqrt{s_1} \cdot \left\|\mathbf{W}^{(t+0.5)} - \overline{\mathbf{W}}^{(t+0.5)}\right\|_{\infty,\infty}, \quad (A.19)
\end{aligned}$$

where the equality follows from the definition of $\mathrm{HT}(\cdot, \cdot)$ function and the first inequality holds due to the sparsity property of $\mathbf{W}^{(t+0.5)}$ and $\overline{\mathbf{W}}^{(t+0.5)}$. Recall that from Algorithm 1 and (5.5), we know that:

$$\begin{aligned}
\mathbf{W}^{(t+0.5)} &= \mathbf{W}^{(t)} - \eta_1 \nabla_1 f_n(\mathbf{W}^{(t)}, \mathbf{\Omega}^{(t)}), \\
\overline{\mathbf{W}}^{(t+0.5)} &= M_1(\mathbf{W}^{(t)}, \mathbf{\Omega}^{(t)}) = \mathbf{W}^{(t)} - \eta_1 \nabla_1 f(\mathbf{W}^{(t)}, \mathbf{\Omega}^{(t)}).
\end{aligned}$$

Let $\eta_1 \leq 2\nu\tau/(2\nu^2\tau + 1)$, thus (A.19) can be further bounded by

$$\begin{aligned}
&\left\|\mathrm{HT}(\mathbf{W}^{(t+0.5)}, \widehat{S}_1^{(t+0.5)}) - \mathrm{HT}(\overline{\mathbf{W}}^{(t+0.5)}, \widehat{S}_1^{(t+0.5)})\right\|_F \\
&\leq \frac{2\nu\tau}{2\nu^2\tau + 1}\sqrt{s_1} \cdot \left\|\nabla_1 f_n(\mathbf{W}^{(t)}, \mathbf{\Omega}^{(t)}) - \nabla_1 f(\mathbf{W}^{(t)}, \mathbf{\Omega}^{(t)})\right\|_{\infty,\infty} \\
&\leq \frac{2\nu\tau}{2\nu^2\tau + 1}\sqrt{s_1} \cdot \epsilon_1,
\end{aligned}$$

where the last inequality follows from Assumption 5.4. Similarly, we can build the bound for (5.10):

$$\left\|\mathrm{HT}(\mathbf{\Omega}^{(t+0.5)}, \widehat{S}_2^{(t+0.5)}) - \mathrm{HT}(\overline{\mathbf{\Omega}}^{(t+0.5)}, \widehat{S}_2^{(t+0.5)})\right\|_F \leq \frac{1568R^2}{2401R^4 + 256}\sqrt{s_2} \cdot \epsilon_2.$$

□

## A.8 Proof of Lemma 5.9

In order to prove Lemma 5.9, we need the following auxiliary lemma.

**Lemma A.2.** Under Assumption 4.2, if $\mathbf{W}$ is a $s_1^*$-sparse matrix, we have with probability at least $1 - 1/n^2$ that

$$\frac{1}{n}\|\mathbf{X}\mathbf{W}\|_F^2 \geq \frac{1}{2\tau}\|\mathbf{W}\|_F^2,$$

when the sample size $n$ is sufficiently large.

*Proof of Lemma 5.9.* For the multivariate Lasso problem in Algorithm 2:

$$\min_{\mathbf{W}} \left\{ \frac{1}{2n}\|\mathbf{Y} - \mathbf{X}\mathbf{W}\|_F^2 + \lambda\|\mathbf{W}\|_{1,1} \right\}, \tag{A.20}$$

it is equivalent to

$$\min_{\mathbf{W}} \left\{ \frac{1}{2n}\|\mathrm{vec}(\mathbf{Y}) - \mathrm{vec}(\mathbf{X}\mathbf{W})\|_F^2 + \lambda\|\mathrm{vec}(\mathbf{W})\|_1 \right\}. \tag{A.21}$$

By the property of Kronecker product that $\mathrm{vec}(\mathbf{ABC}) = (\mathbf{C}^\top \otimes \mathbf{A})\mathrm{vec}(\mathbf{B})$, we obtain

$$\mathrm{vec}(\mathbf{X}\mathbf{W}) = \mathrm{vec}(\mathbf{X}\mathbf{W}\mathbf{I}) = (\mathbf{I} \otimes \mathbf{X})\mathrm{vec}(\mathbf{W}).$$

Submit this result into (A.21), we get

$$\min_{\mathbf{W}} \left\{ \frac{1}{2n}\|\mathrm{vec}(\mathbf{Y}) - (\mathbf{I} \otimes \mathbf{X})\mathrm{vec}(\mathbf{W})\|_F^2 + \lambda\|\mathrm{vec}(\mathbf{W})\|_1 \right\}.$$

Thus we transform (A.20) into a standard Lasso problem where $\mathrm{vec}(\mathbf{Y})$ is the response vector, $(\mathbf{I} \otimes \mathbf{X})$ is the design matrix, and $\mathrm{vec}(\mathbf{W})$ is the regression coefficient vector. By Corollary 2 in Negahban et al. (2009), we have with probability at least $1 - 1/(dm)$ that

$$\|\widehat{\mathbf{W}} - \mathbf{W}^*\|_F \leq C'\nu\tau\sqrt{\frac{s_1^* \cdot \log(dm)}{n}},$$

provided that the following holds

$$\frac{1}{n}\|(\mathbf{I} \otimes \mathbf{X})\mathrm{vec}(\mathbf{W})\|_F^2 \geq \frac{1}{2\tau}\|\mathrm{vec}(\mathbf{W})\|_2^2. \tag{A.22}$$

The remaining task is to verify (A.22). According to Lemma A.2, when $n$ is sufficiently large, with probability at least $1 - 1/n^2$ we have

$$\frac{1}{n}\|\mathbf{X}\mathbf{W}\|_F^2 \geq \frac{1}{2\tau}\|\mathbf{W}\|_F^2,$$

which leads to

$$\frac{1}{2\tau}\|\mathrm{vec}(\mathbf{W})\|_2^2 \leq \frac{1}{n}\|\mathrm{vec}(\mathbf{X}\mathbf{W})\|_F^2 = \frac{1}{n}\|(\mathbf{I} \otimes \mathbf{X})\mathrm{vec}(\mathbf{W})\|_F^2.$$

Thus (A.22) is satisfied. According to Algorithm 2, $\mathbf{W}^{\mathrm{init}} = \widehat{\mathbf{W}}$, and the proof is completed. $\square$

## A.9 Proof of Lemma 5.10

In order to prove Lemma 5.10, we need the following auxiliary lemma and theorem.

**Lemma A.3.** *Under Assumptions 4.1 and 4.2, with probability at least $1 - 1/d - 4/n^2$, the covariance matrix $\mathbf{S}$ in Algorithm 2 satisfies*

$$\|\mathbf{S} - \mathbf{\Sigma}^*\|_{\infty,\infty} \leq C\nu\sqrt{\frac{\log m}{n}} + C'\nu^2\tau^2 s_1^* \cdot \frac{\log m + \log d}{n}.$$

**Theorem A.4** (Loh and Wainwright (2013))**.** *Under Assumptions 4.1 and 4.2, let $\widehat{\mathbf{\Omega}}$ be the graphical Lasso estimator, $\mathbf{S}$ be the sample covariance matrix and $\mathbf{\Sigma}^*$ be the true covariance matrix. $\mathbf{\Omega}^* = \mathbf{\Sigma}^{*-1}$ is a $s_2^*$-sparse precision matrix. If $\lambda \geq \|\mathbf{S} - \mathbf{\Sigma}^*\|_{\infty,\infty}$, then we have*

$$\|\widehat{\mathbf{\Omega}} - \mathbf{\Omega}^*\|_F \leq C'\lambda\nu^2\sqrt{s_2^*}.$$

*Proof of Lemma 5.10.* By combining Lemma A.3 and Theorem A.4, we have

$$\|\widehat{\mathbf{\Omega}} - \mathbf{\Omega}^*\|_F \leq C'\nu^3\sqrt{\frac{s_2^* \cdot \log m}{n}} + C''\nu^4\tau^2 s_1^* \sqrt{s_2^*} \cdot \frac{\log m + \log d}{n}.$$

According to Algorithm 2, $\mathbf{\Omega}^{\text{init}} = \widehat{\mathbf{\Omega}}$, and the proof is completed. □

# B Proof of Auxiliary Lemmas in Section A

## B.1 Proof of Lemma A.2

*Proof.* We have

$$\frac{1}{n}\|\mathbf{X}\mathbf{W}\|_F^2 = \text{tr}(\mathbf{W}^\top \widehat{\mathbf{\Sigma}}_X \mathbf{W}) = \text{vec}(\mathbf{W})^\top \text{vec}(\widehat{\mathbf{\Sigma}}_X \mathbf{W}\mathbf{I})$$
$$= \text{vec}(\mathbf{W})^\top (\mathbf{I} \otimes \widehat{\mathbf{\Sigma}}_X) \text{vec}(\mathbf{W})$$
$$= \text{vec}(\mathbf{W})^\top (\mathbf{I} \otimes \widehat{\mathbf{\Sigma}}_X - \mathbf{I} \otimes \mathbf{\Sigma}_X) \text{vec}(\mathbf{W}) + \text{vec}(\mathbf{W})^\top (\mathbf{I} \otimes \mathbf{\Sigma}_X) \text{vec}(\mathbf{W}),$$

where the second equality holds due to that $\text{tr}(\mathbf{A}^\top \mathbf{B}) = \text{vec}(\mathbf{A})^\top \text{vec}(\mathbf{B})$, the third equality follows from that $\text{vec}(\mathbf{ABC}) = (\mathbf{C}^\top \otimes \mathbf{A})\text{vec}(\mathbf{B})$. By Assumption 4.2, $\lambda_{\min}(\mathbf{\Sigma}_X) = 1/\tau$, thus the above equality can be bounded by

$$\frac{1}{n}\|\mathbf{X}\mathbf{W}\|_F^2 \geq \lambda_{\min}(\mathbf{I} \otimes \mathbf{\Sigma}_X) \cdot \|\mathbf{W}\|_F^2 - \left|\text{vec}(\mathbf{W})^\top \cdot (\mathbf{I} \otimes (\widehat{\mathbf{\Sigma}}_X - \mathbf{\Sigma}_X)) \cdot \text{vec}(\mathbf{W})\right|. \quad \text{(B.1)}$$

Since $\text{vec}(\mathbf{W})$ has the same sparsity level $s_1^*$ as $\mathbf{W}$ does, suppose the support set for $\text{vec}(\mathbf{W})$ is $S$, (B.1) can be further bounded by

$$\frac{1}{n}\|\mathbf{X}\mathbf{W}\|_F^2 \geq \frac{1}{\tau}\|\mathbf{W}\|_F^2 - \left\|(\mathbf{I} \otimes (\widehat{\mathbf{\Sigma}}_X - \mathbf{\Sigma}_X))_{SS}\right\|_2 \cdot \|\mathbf{W}\|_F^2$$
$$\geq \frac{1}{\tau}\|\mathbf{W}\|_F^2 - C\lambda_{\max}((\mathbf{\Sigma}_X)_{SS})\sqrt{\frac{s_1^*}{n}}\|\mathbf{W}\|_F^2, \quad \text{(B.2)}$$

where the last inequality holds due to Lemma C.6. Thus when the sample size $n$ is sufficiently large, with probability at least $1 - 1/n^2$ we have $\|\mathbf{X}\mathbf{W}\|_F^2/n$ greater than $\|\mathbf{W}\|_F^2/2\tau$. This completes the proof. □

## B.2 Proof of Lemma A.3

*Proof.* Since we have

$$\mathbf{S} = \frac{1}{n}(\mathbf{Y} - \mathbf{X}\widehat{\mathbf{W}})^\top(\mathbf{Y} - \mathbf{X}\widehat{\mathbf{W}})$$

$$= \frac{1}{n}(\mathbf{Y} - \mathbf{X}\mathbf{W}^* + \mathbf{X}\mathbf{W}^* - \mathbf{X}\widehat{\mathbf{W}})^\top(\mathbf{Y} - \mathbf{X}\mathbf{W}^* + \mathbf{X}\mathbf{W}^* - \mathbf{X}\widehat{\mathbf{W}})$$

$$= \frac{1}{n}(\mathbf{Y} - \mathbf{X}\mathbf{W}^*)^\top(\mathbf{Y} - \mathbf{X}\mathbf{W}^*) + \frac{1}{n}(\widehat{\mathbf{W}} - \mathbf{W}^*)^\top \mathbf{X}^\top \mathbf{X} (\widehat{\mathbf{W}} - \mathbf{W}^*) + \frac{2}{n}(\mathbf{Y} - \mathbf{X}\mathbf{W}^*)^\top \mathbf{X}(\mathbf{W}^* - \widehat{\mathbf{W}}).$$

Thus,

$$\|\mathbf{S} - \mathbf{\Sigma}^*\|_{\infty,\infty}$$
$$\leq \underbrace{\|\widehat{\mathbf{\Sigma}} - \mathbf{\Sigma}^*\|_{\infty,\infty}}_{I_1} + \underbrace{\frac{1}{n}\|(\widehat{\mathbf{W}} - \mathbf{W}^*)^\top \mathbf{X}^\top \mathbf{X}(\widehat{\mathbf{W}} - \mathbf{W}^*)\|_{\infty,\infty}}_{I_2} + \underbrace{\frac{2}{n}\|(\mathbf{Y} - \mathbf{X}\mathbf{W}^*)^\top \mathbf{X}(\mathbf{W}^* - \widehat{\mathbf{W}})\|_{\infty,\infty}}_{I_3}.$$

(B.3)

For term $I_1$, by Lemma C.7, we have

$$\|\widehat{\mathbf{\Sigma}} - \mathbf{\Sigma}^*\|_{\infty,\infty} \leq C\lambda_{\max}(\mathbf{\Sigma}^*)\sqrt{\frac{\log m}{n}} = C\nu\sqrt{\frac{\log m}{n}}.$$

For term $I_2$, we have

$$\frac{1}{n}\|(\widehat{\mathbf{W}} - \mathbf{W}^*)^\top \mathbf{X}^\top \mathbf{X}(\widehat{\mathbf{W}} - \mathbf{W}^*)\|_{\infty,\infty} \leq \frac{1}{n}\|(\widehat{\mathbf{W}} - \mathbf{W}^*)^\top \mathbf{X}^\top \mathbf{X}(\widehat{\mathbf{W}} - \mathbf{W}^*)\|_2$$
$$\leq \frac{1}{n}\|\mathbf{X}^\top \mathbf{X}\|_2 \cdot \|\widehat{\mathbf{W}} - \mathbf{W}^*\|_F^2$$
$$\leq \frac{C''\nu^2\tau^2 s_1^* \cdot \log(dm)}{n},$$

where the first and second inequalities hold due to the matrix norm inequalities and the third inequality follows from Assumption 4.2 and also the conclusion from Theorem 5.9.

For term $I_3$, we have

$$\frac{2}{n}\|(\mathbf{Y} - \mathbf{X}\mathbf{W}^*)^\top \mathbf{X}(\mathbf{W}^* - \widehat{\mathbf{W}})\|_{\infty,\infty} \leq \frac{2}{n}\|(\mathbf{Y} - \mathbf{X}\mathbf{W}^*)^\top \mathbf{X}\|_{\infty,\infty} \cdot \|\mathbf{W}^* - \widehat{\mathbf{W}}\|_{1,1}$$
$$\leq \left\|\frac{2}{n}\sum_{i=1}^n \boldsymbol{\epsilon}_i \mathbf{x}_i^\top\right\|_{\infty,\infty} \cdot \sqrt{s_1^* + s_1} \cdot \|\widehat{\mathbf{W}} - \mathbf{W}^*\|_F, \quad \text{(B.4)}$$

where the inequalities follow from the matrix norm inequalities. Note that for $\boldsymbol{\epsilon}_i \mathbf{x}_i^\top$ we have

$$\big\|[\boldsymbol{\epsilon}_i \mathbf{x}_i^\top]_{jk}\big\|_{\psi_2}^2 = \big\|x_{ij}\epsilon_{ik}\big\|_{\psi_2}^2 \leq \big\|\epsilon_{ik}\big\|_{\psi_2}^2 \leq C_1^2\nu^2.$$

By Hoeffding-type inequality in Theorem C.4, we have

$$\mathbb{P}\left\{\left|\frac{1}{n}\sum_{i=1}^n [\boldsymbol{\epsilon}_i \mathbf{x}_i^\top]_{jk}\right| \geq t\right\} \leq \exp\left(-\frac{C_8 t^2 n}{C_1^2 \nu^2}\right).$$

Apply union bound to all possible pairs of $j \in [m]$, $k \in [d]$, we get

$$\mathbb{P}\bigg\{\bigg\|\frac{1}{n}\sum_{i=1}^{n}\boldsymbol{\epsilon}_i\mathbf{x}_i^\top\bigg\|_{\infty,\infty} \geq t\bigg\} \leq md\exp\bigg(-\frac{C_8 t^2 n}{C_1^2 \nu^2}\bigg).$$

Choose $t = \nu C_9 \sqrt{\log(md)/n}$ and $C_9 = C_1/\sqrt{C_8}$ with probability at least $1 - 1/d$ we have that

$$\bigg\|\frac{1}{n}\sum_{i=1}^{n}\boldsymbol{\epsilon}_i\mathbf{x}_i^\top\bigg\|_{\infty,\infty} \leq C_9 \nu \sqrt{\frac{\log m + \log d}{n}}. \tag{B.5}$$

Submit the above result and the conclusion from Theorem 5.9 into (B.4), we have

$$\frac{2}{n}\|(\mathbf{Y}-\mathbf{X}\mathbf{W}^*)^\top \mathbf{X}(\mathbf{W}^* - \widehat{\mathbf{W}})\|_{\infty,\infty} \leq C'''\nu^2\tau\sqrt{s_1^*(s_1^* + s_1)} \cdot \frac{\log m + \log d}{n}. \tag{B.6}$$

Combine the results for term $I_1, I_2, I_3$, we obtain

$$\|\mathbf{S} - \boldsymbol{\Sigma}^*\|_{\infty,\infty} \leq C\nu\sqrt{\frac{\log m}{n}} + \bigg(C''\nu^2\tau^2 s_1^* + C'''\nu^2\tau\sqrt{s_1^*(s_1^* + s_1)}\bigg) \cdot \frac{\log m + \log d}{n}$$

$$\leq C\nu\sqrt{\frac{\log m}{n}} + C'\nu^2\tau^2 s_1^* \cdot \frac{\log m + \log d}{n}. \tag{B.7}$$

$\square$

## C  Additional Auxiliary Lemmas

**Lemma C.1** (Rotation invariance Vershynin (2010))**.** For a set of independent centered sub-gaussian random variables $X_i$, $\sum_i a_i \|X_i\|_{\nu_2}^2$ is also a centered sub-gaussian random variable, and further, we have

$$\bigg\|\sum_i a_i X_i\bigg\|_{\psi_2}^2 \leq C \sum_i a_i^2 \|X_i\|_{\psi_2}^2,$$

where $C$ is an absolute constant.

**Lemma C.2** (Product Property)**.** For any two sub-gaussian random variables $X$ and $Y$, we have

$$\|XY\|_{\psi_1} \leq 2\|X\|_{\psi_2} \cdot \|Y\|_{\psi_2}.$$

**Lemma C.3** (Centering Vershynin (2010))**.** For any sub-exponential random variables $X$, we have

$$\|X - \mathbb{E}X\|_{\psi_1} \leq 2\|X\|_{\psi_1}.$$

**Theorem C.4** (Proposition 5.10 in Vershynin (2010))**.** Let $X_1, X_2, \ldots, X_n$ be independent centered sub-Gaussian random variables, and let $K = \max_i \|X_i\|_{\psi_2}$. Then for every $a = (a_1, a_2, \ldots, a_n) \in \mathbb{R}^n$ and for every $t > 0$, we have

$$\mathbb{P}\bigg(\bigg|\sum_{i=1}^{n} a_i X_i\bigg| > t\bigg) \leq \exp\bigg(-\frac{Ct^2}{K^2\|a\|_2^2}\bigg),$$

where $C > 0$ is a constant.

**Theorem C.5** (Proposition 5.16 in Vershynin (2010)). *Let $X_1, X_2, \ldots, X_n$ be independent centered sub-exponential random variables, and let $K = \max_i \|X_i\|_{\psi_1}$. Then for every $a = (a_1, a_2, \ldots, a_n) \in \mathbb{R}^n$ and for every $t > 0$, we have*

$$\mathbb{P}\bigg(\bigg|\sum_{i=1}^n a_i X_i\bigg| > t\bigg) \leq 2\exp\bigg[-C\min\bigg(\frac{t^2}{K^2\|a\|_2^2}, \frac{t}{K\|a\|_\infty}\bigg)\bigg],$$

*where $C > 0$ is a constant.*

**Lemma C.6** (Vershynin (2010)). *Suppose $S$ is a support set with $|S| = s$, we have with probability at least $1 - 1/n^2$ that*

$$\|\widehat{\boldsymbol{\Sigma}}_{SS} - \boldsymbol{\Sigma}_{SS}\|_2 \leq C\lambda_{\max}(\boldsymbol{\Sigma}_{SS})\sqrt{\frac{s}{n}},$$

*where $C$ is some universal constant.*

**Lemma C.7** (Loh and Wainwright (2013)). *Suppose $\widehat{\boldsymbol{\Sigma}}$ is the solution for the glasso problem in Algorithm 2 and $\boldsymbol{\Sigma}^*$ is the minimizer of the glasso problem, then with probability at least $1 - 4/n^2$, we have*

$$\|\widehat{\boldsymbol{\Sigma}} - \boldsymbol{\Sigma}^*\|_{\infty,\infty} \leq C\lambda_{\max}(\boldsymbol{\Sigma}^*)\sqrt{\frac{\log m}{n}}.$$